\documentclass[11pt]{article}

\usepackage[preprint]{acl}

\usepackage{times}
\usepackage{latexsym}

\usepackage[T1]{fontenc}

\usepackage[utf8]{inputenc}

\usepackage{microtype}

\usepackage{inconsolata}

\usepackage{graphicx}
\usepackage{subcaption}
\usepackage{booktabs}
\usepackage{amsmath}
\usepackage[dvipsnames]{xcolor}

\usepackage{CJKutf8}
\usepackage{tcolorbox}
\tcbuselibrary{breakable}
\usepackage{fancyvrb}
\usepackage{multirow}

\usepackage{booktabs}
\usepackage{makecell}
\usepackage[normalem]{ulem} 

\usepackage{twemojis}

\newcommand{\gold}{\twemoji{1f947}}
\newcommand{\silver}{\twemoji{1f948}}
\newcommand{\bronze}{\twemoji{1f949}}

\title{Dango: A Strictly L1-Only Large Language Model \\ for Studying Second Language Acquisition}

\author{%
  Shiho Matta${}^{1}$ Yin Jou Huang${}^{1}$ Fei Cheng${}^{1}$ Takashi Kodama${}^{2}$\\
 \textbf{Hirokazu Kiyomaru}${}^{2}$ \textbf{Yugo Murawaki}${}^{1}$\\
${}^{1}$Kyoto University ${}^{2}$NII-LLMC\\
\texttt{\{matta,huang,feicheng,murawaki\}@nlp.ist.i.kyoto-u.ac.jp}\\
\texttt{\{tkodama,kiyomaru\}@nii.ac.jp}\\
}

\begin{document}
\maketitle
\begin{abstract}

We introduce \textbf{Dango}, a 1.8B-parameter large language model designed for controlled studies of L1→L2 (Japanese→English) transfer in second language acquisition (SLA). While previous studies have explored SLA in language models, they have predominantly relied on smaller or non-decoder models, limiting their ability to generate open-ended text and reducing their suitability as practical L2 simulators. We identify a key challenge when scaling models to this size: L2 contamination within the "monolingual" pretraining corpus used for L1 acquisition. To address this, we propose a filtering method to reduce premature exposure to English while preserving realistic, minimal exposure. We then fine-tune the model on LLM-generated L2-learning lessons to simulate the L2 acquisition process. Our evaluations confirm that Dango develops human-like L2 production patterns, outperforming both unfiltered and standard multilingual baselines. We release the model, data, and code to facilitate reproducible computational SLA research and learner-facing applications.\footnote{We will soon release the model, data, and code.}

\end{abstract}

\section{Introduction}

Second language acquisition (SLA) refers to the learning of a second language (L2) after a first language (L1) has already been acquired natively. In humans, 
the influence of L1 can facilitate L2 learning in some cases and interfere with it in others, a phenomenon widely known as the language transfer effect \cite{lado1957linguistics, odlin1989language}. This may result in overuse, avoidance, and misuse patterns of certain linguistic features in a speaker's L2 production \cite{Jarvis2008}. For example, because Japanese lacks a grammatical article system, Japanese-speaking learners of English often show difficulty with English article use \cite{murakami_2015_l1}. This can lead to article-omission errors such as ``I went to library'' instead of ``I went to the library.''

\begin{figure}[t]
    \centering

    \includegraphics[width=0.99\linewidth]{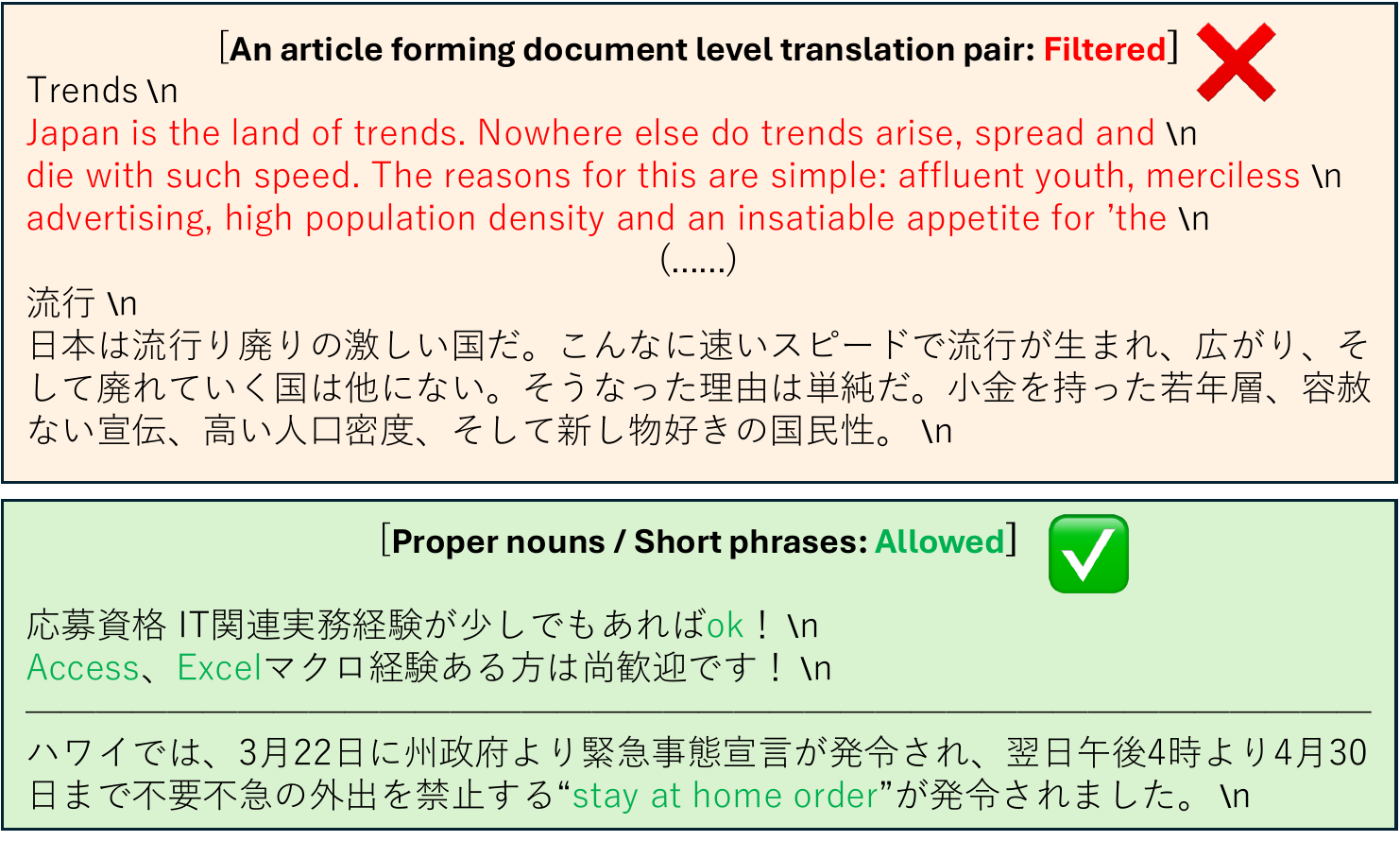}

    \vspace{-2mm}
    \caption{English contamination examples: severe case of translation pairs versus minor, typical English exposure for Japanese speakers.}
    \label{fig:fig_l2_contamination_example}
    \vspace{-5mm}
\end{figure}

As transformer-based language models (LMs) have recently shown strong performance on many linguistic tasks \cite{lai-etal-2023-chatgpt}, many studies have investigated whether they also exhibit such human-like transfer effects \cite{oba-etal-2023-second, aoyama-schneider-2024-modeling, yadavalli-etal-2023-slabert, constantinescu-etal-2025-investigating}. 
These studies train LMs from scratch first on L1 and then on L2, thereby mimicking the order in which humans learn languages. Transfer effects are then evaluated, for example, by testing the model’s grammatical judgments on the L2 and comparing them with human tendencies \cite{yadavalli-etal-2023-slabert}.

Although these studies provide valuable insights into transfer effects in LMs, their use as practical simulators of L2 speakers remains limited. This is largely due to limitations of architecture and scale. Prior work has used either encoder-only models \cite{oba-etal-2023-second, yadavalli-etal-2023-slabert, constantinescu-etal-2025-investigating} or small decoder-only models, such as the 137M-parameter GPT-2 in \citet{constantinescu-etal-2025-investigating}. As a result, these models are less suitable for robust open-ended or interactive generation, and for potential applications such as L2 learner support \cite{lyu_2024_effectiveness} and teacher training with virtual L2 speakers \cite{vasmatzoglou-nichiarain-2020-gamebased-simulation-training-english-teacher}.


To equip future SLA research with a more modern and capable model, and to enable broader practical applications, we release \textbf{Dango}, a Llama-2-like decoder-only transformer trained in an L1→L2 manner at a much larger scale. Dango has 1.8B parameters and is pretrained on far more data than prior work, with up to 100B L1-stage tokens. 







\begin{figure*}[!ht] 
  \centering
  \includegraphics[width=0.95\textwidth]{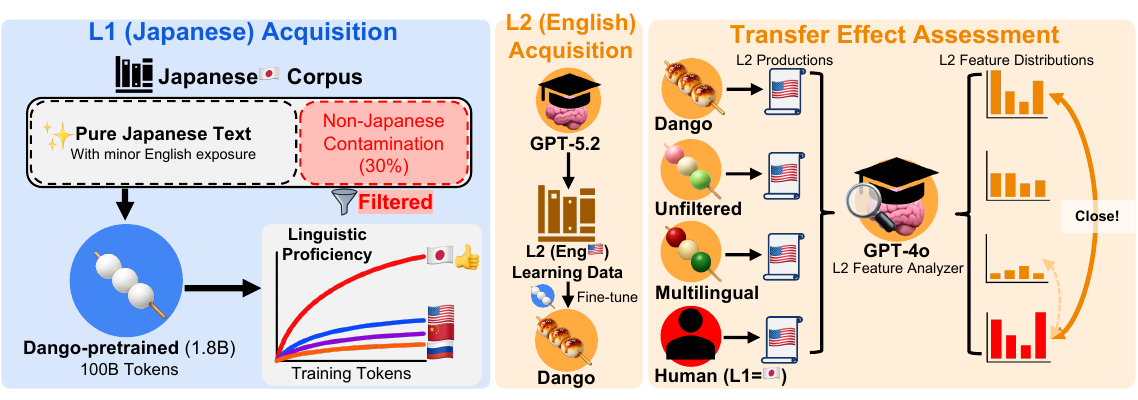}
  \vspace{-3mm}
  \caption{An overview of our proposed methodology.}
  \label{fig:overview}
  \vspace{-5mm}
\end{figure*}

To train an LLM at this size, web-crawled text is the only viable data source \cite{hoffmann2022trainingcomputeoptimallargelanguage_chinchilla}. 
However, the Japanese web corpus we used for L1 acquisition contained substantial accidental English exposure. We call this problem \textbf{L2 contamination}. Although Japanese speakers naturally see English through loanwords, proper nouns, and code-switching \cite{hyde_2002_japans, Japanese-English}, systematic acquisition of English is relatively rare outside classroom-like educational settings \cite{barrs_2020_learning}. 
It is therefore natural to reshape the corpus into a state closer to a typical Japanese speaker’s environment: removing structured English content, such as in the worst case, translation pairs
, while retaining minimal English such as brand names and simple phrases (Figure~\ref{fig:fig_l2_contamination_example}). We designed a filtering pipeline that removes excessive English exposure, reducing the corpus size by 30\% (Figure~\ref{fig:overview}).
In the L2 acquisition stage, we fine-tune Dango on LLM-synthesized textbook-like lessons. The lessons follow a gradual difficulty progression, as they are based on an English seed word drawn from a word list with difficulty labels. The lessons are anchored in both Japanese and English, containing (i) a Japanese explanation of the seed word, (ii) an example English sentence using the word, paired with a Japanese translation, and (iii) a brief Japanese commentary highlighting relevant grammar or usage. In this way, Dango is exposed to English while keeping instruction and metalinguistic reasoning in its L1, which mirrors common textbook learning practice \cite{nakayama2022close}.



Our experiments show that Dango acquires English generation abilities, including the capacity to produce long English sentences absent from its strictly filtered pretraining data. Evaluations further show that, among all methods, Dango’s English outputs most closely resemble those of human Japanese learners of English, outperforming unfiltered and multilingual baselines and matching prompted GPT-5.5. These results provide evidence of successful language transfer. We release Dango’s code and checkpoints to support controlled SLA research and practical applications.




\section{Related Work}

\subsection{Second Language Acquisition (SLA) with Language Models (LMs)}
\paragraph{Modeling SLA by Training in an L1→L2 Setup. } Previous studies have trained LMs on L1 and L2 sequentially to model human L2 acquisition. 
\citet{oba-etal-2023-second} trained 18M-parameter encoder-only models and analyzed how different L1 pretraining languages shape the trajectory and ultimate success of L2 grammatical generalization. \citet{yadavalli-etal-2023-slabert} also trained 110M-parameter encoder-only models, but aimed to isolate positive and negative cross-linguistic transfer. \citet{aoyama-schneider-2024-modeling} trained 11M-parameter (our estimation) GPT-2 models as L2 English speakers, investigating transfer effects and showing that LM surprisal helps predict L2 English readers’ word-by-word reading times. \citet{constantinescu-etal-2025-investigating} trained 125M-parameter encoder-only and 137M-parameter decoder-only models, but their focus was on recreating the critical period effect observed in humans rather than directly examining transfer effects.

\paragraph{Training Data.} Training data is central to modeling SLA. Some work has attempted to create a more human-like setting by using data derived from human speech. For example, \citet{yadavalli-etal-2023-slabert} used child-directed speech, and \citet{constantinescu-etal-2025-investigating} partially used spoken-language data. However, these data sources are limited in size, and \citet{constantinescu-etal-2025-investigating} had to include web text to obtain sufficient data and cross-linguistic coverage. On the other hand, \citet{oba-etal-2023-second} and \citet{aoyama-schneider-2024-modeling} relied on CC-100 \cite{conneau-etal-2020-unsupervised, wenzek-etal-2020-ccnet} web-crawled data for L1 training.

\paragraph{Evaluating Transfer Effect.}
\citet{oba-etal-2023-second}, \citet{aoyama-schneider-2024-modeling}, \citet{yadavalli-etal-2023-slabert}, and \citet{constantinescu-etal-2025-investigating} used BLiMP \cite{warstadt-etal-2020-blimp-benchmark} to measure the L2 (English) grammatical generalization. 
Other probes include perplexity \cite{aoyama-schneider-2024-modeling, constantinescu-etal-2025-investigating}, model surprisal \cite{aoyama-schneider-2024-modeling} and knowledge-based multiple-choice tasks \cite{constantinescu-etal-2025-investigating}. 

However, because these probes rely entirely on the predefined inputs within existing benchmarks, they do not truly test a model's generation ability. To address this limitation, we evaluate the transfer effect in Dango's English text production using the automatic LLM-as-a-judge framework introduced by \citet{gao-etal-2025-llms}. We directly annotate linguistic features in the model's output to compare their distribution with authentic human L2 productions.

\subsection{Controlling Knowledge Exposure during Pretraining}
\paragraph{LLMs with Knowledge Cutoff.}
Recent efforts highlight the necessity of strictly constrained pretraining environments to isolate specific phenomena from the massive data contamination inherent in standard LLMs. For instance, TimeCapsule LLM is trained exclusively on texts from constrained historical periods to accurately emulate the vocabulary and worldview of that era without the intrusion of modern concepts.\footnote{\url{https://github.com/haykgrigo3/TimeCapsuleLLM}} Similarly, History LLMs are trained from scratch using strict temporal knowledge cutoffs to completely prevent the leakage of future knowledge.\footnote{\url{https://github.com/DGoettlich/history-llms}} These projects demonstrate that true behavioral simulation requires a verifiable, uncontaminated training baseline.

\paragraph{L2 Contamination in Web Corpora.}
L2 contamination matters in controlled studies of second-language acquisition and transfer effects. Because language identification is imperfect, corpora intended to be non-English often still contain English data \cite{caswell-etal-2020-language}. In controlled experiments, such accidental exposure can be problematic because it may introduce unintended cross-linguistic transfer \cite{blevins-zettlemoyer-2022-language}. This issue may also apply to prior work that used web corpora for training \cite{oba-etal-2023-second, aoyama-schneider-2024-modeling, constantinescu-etal-2025-investigating}, although any resulting failures may have been too subtle to detect at their scale. To our knowledge, our work is the first to directly address the L2 contamination problem in computational SLA.



\section{Methodology}
The training of Dango consists of two stages (Figure~\ref{fig:overview}). First, we model L1 (Japanese) acquisition by pretraining a decoder-only language model on a strictly filtered Japanese corpus (\S~\ref{sec:3_l1_acquisition}), which yields the \textbf{Dango-pretrained} model. 
Second, we model L2 (English) acquisition by fine-tuning the pretrained model on LLM-generated L2 learning data, yielding \textbf{Dango} (\S~\ref{sec:3_l2_acquisition}). 

\subsection{L1 Acquisition}
\label{sec:3_l1_acquisition}
We conducted filtering to remove non-Japanese exposures in a Japanese corpus and trained a decoder-only transformer model on the resulting data. 
\subsubsection{Corpus for L1 Acquisition and Filtering}
\label{sec:3_l1_corpus_and_filtering}
To bootstrap our study from an established baseline, we adopted llm-jp-corpus-v3, the training corpus released by the llm-jp project.\footnote{\url{https://gitlab.llm-jp.nii.ac.jp/datasets/llm-jp-corpus-v3/-/tree/main}} The full corpus is multilingual, containing Japanese, English, Chinese, Korean, and code data. In this work, we used only the Japanese portion, which is mainly composed of Japanese Common Crawl data\footnote{\url{https://commoncrawl.org}.} and PDF data from the National Diet Library’s Web Archiving Project (WARP).\footnote{National Diet Library (Japan). NDL Web Archiving Project (WARP), \url{https://warp.ndl.go.jp/}.} Nevertheless, we observed that this subset still included a non-negligible amount of English text. 

We built a filtering pipeline to limit exposure to English grammar and long, structured English sentences that could confound L1→L2 transfer \cite{barrs_2020_learning, blevins-zettlemoyer-2022-language}, while retaining benign English signals common in Japanese text, such as everyday words, phrases, and proper nouns \cite{hyde_2002_japans, Japanese-English}.

The pipeline combines a whitelist filter with blocklist filters applied at both the document and line levels. The whitelist filter defines the character inventory allowed for Japanese-centric text, including Japanese scripts, common CJK characters, punctuation, symbols, and other frequent marks. Documents with more than 1‰ out-of-inventory characters are discarded, which removes most text in non-CJK and non-Latin scripts, such as Russian and Arabic. Next, blocklist filters remove lines separated by \texttt{``\textbackslash n''} that contain excessive English or Chinese characters. For example, the English filter removes a line if it contains more than four consecutive English words, more than 20 Latin letters, or a Japanese-to-English character ratio below 40\%. If more than 5\% of a document’s lines are removed, the entire document is discarded to avoid damaged context. Further implementation details are provided in Appendix~\ref{appendix:filtering_details}.

\label{method:l1_corpus_and_filtering}

\subsubsection{Pretraining an L1-Only LLM}
\label{sec:3_l1_acquisition_pretraining}

We pretrained a 1.8B-parameter Llama~2--style decoder-only Transformer from scratch on the corpus described in \S~\ref{method:l1_corpus_and_filtering}. To our knowledge, this is the largest model trained from scratch for studying second language acquisition. 
Pretraining details are provided in Appendix~\ref{appendix:pre-training}. 


We adopted the pretrained multilingual tokenizer from the released llm-jp-3 model, which uses a vocabulary of 99,584 tokens covering both Japanese and English. We acknowledge that this gives the model access to L2 subword units from initialization. We leave experiments with more human-like tokenization assumptions to future work.\footnote{\url{https://huggingface.co/llm-jp/llm-jp-3-1.8b}}


\subsubsection{Linguistic Proficiency Assessments.}
\label{sec:3_evaluation}

We monitor the model’s linguistic proficiency during L1 pretraining to assess whether it effectively acquires Japanese and whether unwanted language contamination occurs after filtering.


\paragraph{Assessing Japanese Proficiency.}
We assess Japanese proficiency along two dimensions: grammatical knowledge and general Japanese-language ability. For grammatical knowledge, we use JBLiMP \cite{someya-oseki-2023-jblimp}, a Japanese benchmark in the style of BLiMP \cite{warstadt-etal-2020-blimp-benchmark}. Models are evaluated on minimal pairs that differ in a targeted grammatical property; an item is correct if the model assigns higher probability to the grammatical sentence. Performance is reported as accuracy, with chance at 50\%.

For general Japanese-language ability, we use llm-jp-eval-v1.4.1
,\footnote{\url{https://github.com/llm-jp/scripts/tree/main/evaluation/installers/llm-jp-eval-v1.4.1}} a few-shot benchmark \cite{brown2020languagemodelsfewshotlearners} covering diverse tasks such as natural language inference and multiple-choice question answering. Programming and mathematical tasks are excluded. 


\paragraph{Assessing English Leakage.}
\label{sec:blimp}
We use the English section of MultiBLiMP \cite{jumelet2025multiblimp10massivelymultilingual}, another BLiMP style benchmark.


\paragraph{Assessing Other Leakage.}
We assess Russian and Chinese proficiency using the Russian part in MultiBLiMP and ZhoBLiMP~\cite{liu2025systematicassessmentlanguagemodels_zhoblimp}.

\subsection{L2 Acquisition}
\label{sec:3_l2_acquisition}
To introduce Dango to English, we synthesized English learning data using an off-the-shelf multilingual LLM and fine-tuned the model. 

\subsubsection{Synthesizing L2-Learning Data with LLM}
We generate parallel, lesson-style data for English L2 learning using a strong multilingual LLM. Each example is centered on a seed word from the CEFR-J vocabulary list \cite{tono2019_cefrj_back}, a CEFR-based \cite{coe2001_cefr} inventory tailored to Japanese learners. Since the original CEFR-J annotations cover only A1--B2, we use an externally annotated extension to include words up to C2. We refer to the combined inventory as CEFR-J throughout the paper.\footnote{\url{https://github.com/openlanguageprofiles/olp-en-cefrj/tree/master}
, C1/C2 labels are from the Octanove Vocabulary Profile (CC BY-SA 4.0).}


As shown in Figure~\ref{fig:l2_data_gen_prompt}, we prompt the LLM to generate: (i) a detailed explanation of the seed word, (ii) an example sentence containing the word and its Japanese translation, and (iii) an analysis of salient usage patterns in the example. Since the prompt controls sentence difficulty, average sentence length and complexity increases with the CEFR-J level of the seed word (Figure~\ref{fig:l2_learning_data_sentence_len_dist}). Both the lexical explanation and usage analysis are written in Japanese, reflecting textbook-style English learning for Japanese speakers and encouraging Japanese-to-English transfer.

The above setup yields the \textbf{English--Japanese} (Eng--Jpn) setting, where the model generates an English sentence followed by its Japanese translation. We also create a \textbf{Japanese--English} (Jpn--Eng) variant by simply swapping the English and Japanese sentences in the data. 

\subsubsection{Fine-tuning on L2 Learning Data}
We simulate L2 acquisition by fine-tuning models on L2 learning data. Each training example follows a prompt format in which the user asks for an explanation of a word and the assistant provides the corresponding L2 learning response (Figure~\ref{fig:l2_fine-tuning_prompt}). By combining data from different CEFR levels, we train models that mimic L2 learners at specific proficiency stages (see \S~\ref{sec:5_fine-tuning_on_L2}).

\subsubsection{Assessing L2 Acquisition Progress}
\label{sec:l2_eval_bleu}
We evaluate L2 acquisition progress with BLEU scores on Eng--Jpn and Jpn--Eng translation. We use the translation component of a held-out subset of the L2 learning data for in-distribution testing, and Jpn--Eng parallel sentences from Tatoeba for out-of-distribution evaluation.\footnote{\url{https://tatoeba.org}
. Created by volunteer contributors from the Tatoeba community. Text data licensed under CC BY 2.0 FR. Downloaded 2025-12-05.} 


\subsubsection{L1--L2 Transfer Assessment}
\label{sec:3_transfer_effect_assessment}

To directly quantify L1–L2 transfer effects, we adopt an LLM-as-a-judge evaluation protocol following \citet{gao-etal-2025-llms} to analyze the patterns in the English productions and compare that to human performance. We prompt a judge LLM to extract instances of seven English linguistic feature categories (excluding \textit{Speech Acts}) from model outputs and to assess their contextual correctness: \textit{Number Agreement} (local number concord inside noun phrases), \textit{Tense Agreement} (tense choices consistent with temporal cues), \textit{Subject--Verb Agreement} (verb inflection matching the subject), \textit{Modal Verbs and Expressions}, \textit{Quantifiers and Numerals}, \textit{Noun--Verb Collocations} (conventional verb--noun pairings supporting fluent lexical selection), and \textit{Reference Word} (referential devices such as pronouns/demonstratives that maintain discourse continuity).

As shown in Figure~\ref{fig:annotation_user_prompt_numbers_agreement}, the judge identifies each instance of a target feature in a translation output and returns (i) the exact text span and (ii) a binary label indicating whether the usage is correct in context. From these analysis outputs, we derive for each category (a) a \textbf{Usage Frequency (UF)} distribution (how often the feature appears) and (b) an \textbf{Error Rate (ER)} distribution (how often detected instances are incorrect).

Finally, we examine whether Dango produces English texts that share UF and ER distributions similarly to a human English learner. We apply the same llm-as-a-judge method to extract feature distributions from the ICNALE Written Essays \cite{ishikawa_2013_icnale_written_essay, ishikawa2023icnale_book} corpus. This corpus contains topic-centric essays written by English learners of various L1 backgrounds, including Japanese. We use Jensen--Shannon divergence (JSD) to capture distributional shift in UF, and Bray-Curtis Dissimilarity (BCD) in addition to JSD in ER to better reflect raw error count shifts (Appendix~\ref{appendix:l2_feature_dist_jsd}). We also provide an extended analysis comparing model L2 production against ICNALE Spoken Dialogues \cite{ishikawa_2019_icnale_spoken_dialogue} in  Appendix~\ref{appendix:l2_spoken_dialogues}.



\section{Experiments and Results for L1 (Japanese) Acquisition}
In this section, we describe the experimental setup and evaluation results for pretraining Dango-pretrained, the L1-only LLM. For brevity, we may omit the ``-pretrained'' suffix when referring to the  pretrained models in this section. 






\begin{figure*}[t]
    \centering
    \begin{subfigure}[b]{0.20\linewidth}
        \centering
        \includegraphics[width=\linewidth]{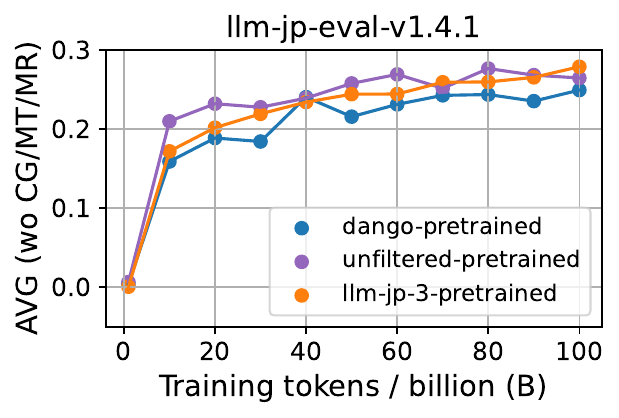}
        \caption{\begin{tabular}[t]{@{}c@{}}Japanese\\(llm-jp-eval-v1.4.1)\end{tabular}}
        \label{fig:l1_llm-jp-eval}
    \end{subfigure}\hfill%
    \begin{subfigure}[b]{0.20\linewidth}
        \centering
        \includegraphics[width=\linewidth]{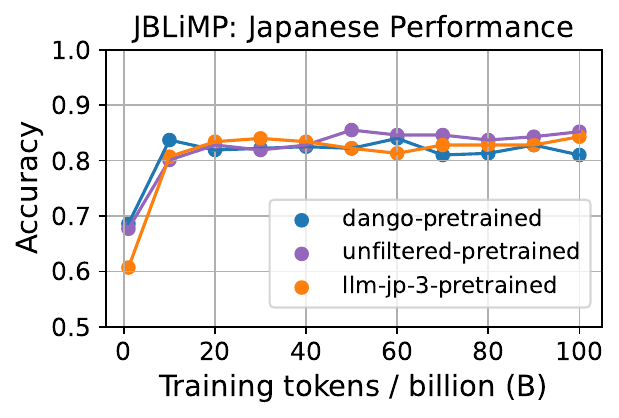}
        \caption{\begin{tabular}[t]{@{}c@{}}Japanese\\(JBLiMP)\end{tabular}}
        \label{fig:l1_blimp_jpn}
    \end{subfigure}\hfill%
    \begin{subfigure}[b]{0.20\linewidth}
        \centering
        \includegraphics[width=\linewidth]{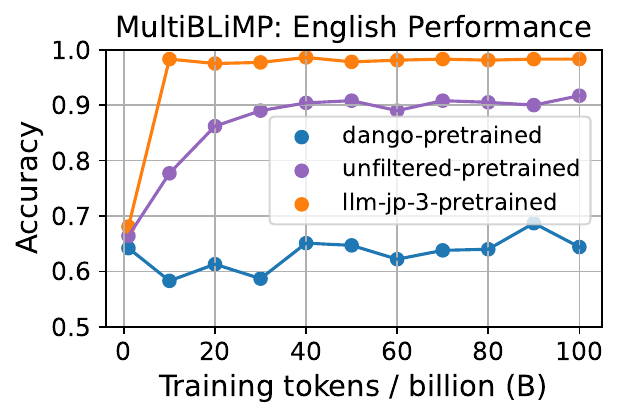}
        \caption{\begin{tabular}[t]{@{}c@{}}English\\(MultiBLiMP)\end{tabular}}
        \label{fig:l1_blimp_eng}
    \end{subfigure}\hfill%
    \begin{subfigure}[b]{0.20\linewidth}
        \centering
        \includegraphics[width=\linewidth]{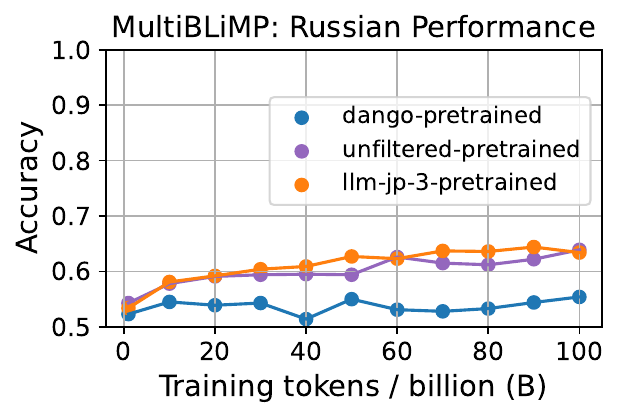}
        \caption{\begin{tabular}[t]{@{}c@{}}Russian\\(MultiBLiMP)\end{tabular}}
        \label{fig:l1_blimp_rus}
    \end{subfigure}\hfill%
    \begin{subfigure}[b]{0.20\linewidth}
        \centering
        \includegraphics[width=\linewidth]{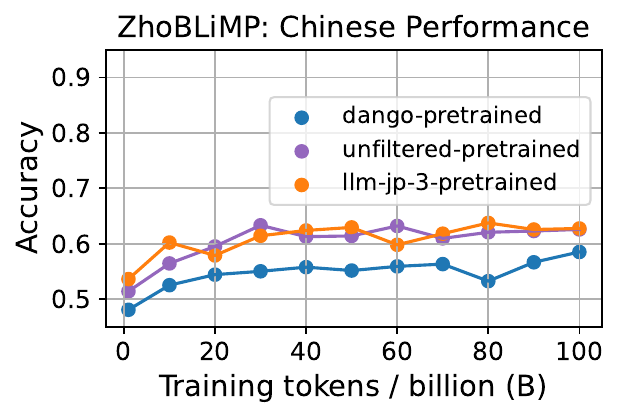}
        \caption{\begin{tabular}[t]{@{}c@{}}Chinese\\(ZhoBLiMP)\end{tabular}}
        \label{fig:l1_blimp_chn}
    \end{subfigure}
    \vspace{-6mm}
    \caption{Linguistic proficiency assessments across the L1-pretraining stage.}
    \vspace{-2mm}
    \label{fig:l1_blimp_result}
\end{figure*}

\subsection{Experimental Settings}
We detail the data preparation and training method for Dango-pretrained and the baselines.

\paragraph{Training-based Baselines. }
For training-based baselines, we trained an \textbf{unfiltered} model and used the \textbf{llm-jp-3} models as multilingual baselines. The unfiltered model was trained on the Japanese portion of llm-jp-corpus-v3 without the filtering described in \S~\ref{sec:3_l1_corpus_and_filtering}. We obtained the llm-jp-3-pretrained checkpoints by contacting the llm-jp team directly. These models have the same size, architecture, and training method as Dango; they differ only in the training data used.

\paragraph{L1 Corpus Filtering and Statistics. }
Detailed filtering parameters are provided in Appendix~\ref{appendix:filtering_details}. Filtering removed 30\% of documents (Table~\ref{tab:l1_corpus_filtered_stats}); the remaining documents were further stripped of English-heavy lines through line-level filtering. Figures~\ref{fig:contamination_document_level_parallel} and~\ref{fig:contamination_word_level_parallel} show document- and phrase-level contamination removed by the English filter. However, short English phrases, such as common daily words and proper nouns, may remain (Figure~\ref{fig:allowed_english}). Thus, Dango can still learn weak word- or phrase-level English associations, but not the ability to generate long, coherent English sentences after L1 pretraining, as shown in Appendix~\ref{appendix:case_study_dango_english_after_L1}.


\paragraph{L1 Pretraining Details. }
Dango-pretrained was trained on 100B tokens from the filtered corpus. Compute-optimal scaling suggests that a 1.8B-parameter model is optimally trained on roughly 35B tokens \cite{hoffmann2022trainingcomputeoptimallargelanguage_chinchilla}; however, we trained for about 3$\times$ more tokens because its Japanese generalization continued improving beyond 35B tokens (Figure~\ref{fig:l1_llm-jp-eval}). 
Additional pretraining details are provided in Appendix~\ref{appendix:pre-training}.

\subsection{Linguistic Proficiency Assessments}
\label{sec:l1_linguistic_proficiency_assessments}

We track the linguistic proficiency of the pretrained models over a 10B-token training interval.\footnote{In Figure~\ref{fig:l1_blimp_result}, the leftmost points on the plot correspond to models trained on 1B tokens.}

\paragraph{Japanese Proficiency. } 
On llm-jp-eval (Figure \ref{fig:l1_llm-jp-eval}), Dango performs slightly below llm-jp-3 as training approaches 100B tokens (-0.029). A category breakdown is provided in Appendix \ref{appendix:llm-jp-eval-radar}. 

On JBLiMP (Figure~\ref{fig:l1_blimp_jpn}), Dango outperforms both baselines in early stage (1B and 10B tokens), likely because it is trained on nearly pure Japanese data. As training continues, the gap closes and all models reach comparable performance. Overall, Dango attains Japanese ability roughly on par with the unfiltered and llm-jp-3 models.

\paragraph{English Proficiency. } On the English portion of MultiBLiMP (Figure \ref{fig:l1_blimp_eng}), llm-jp-3 saturates after only 10B tokens, likely because roughly half of its training tokens are English. In contrast, Dango remains near 0.65 even after 100B tokens. The unfiltered model indicates that naive training on a ``Japanese'' corpus can still yield strong English proficiency, reaching 0.917.

These results suggest that filtering prevents Dango’s English ability from scaling with additional pretraining. The case study in Appendix \ref{appendix:case_study_dango_english_after_L1} further shows that Dango’s English remains largely n-gram-like, likely because the filtering rule removes sequences of more than four consecutive English words. 

\paragraph{Other L2 Leakage.}
As an additional check for L2 leakage, we evaluate Russian and Chinese (Figures \ref{fig:l1_blimp_rus} and \ref{fig:l1_blimp_chn}) to assess the effectiveness of the whitelist and Chinese filter. All three models perform poorly, but Dango shows the lowest proficiency in both languages.




\begin{figure*}[!htp]
  \centering

    \centering
    \begin{subfigure}{0.468\textwidth}
      \centering
      \includegraphics[width=\linewidth]{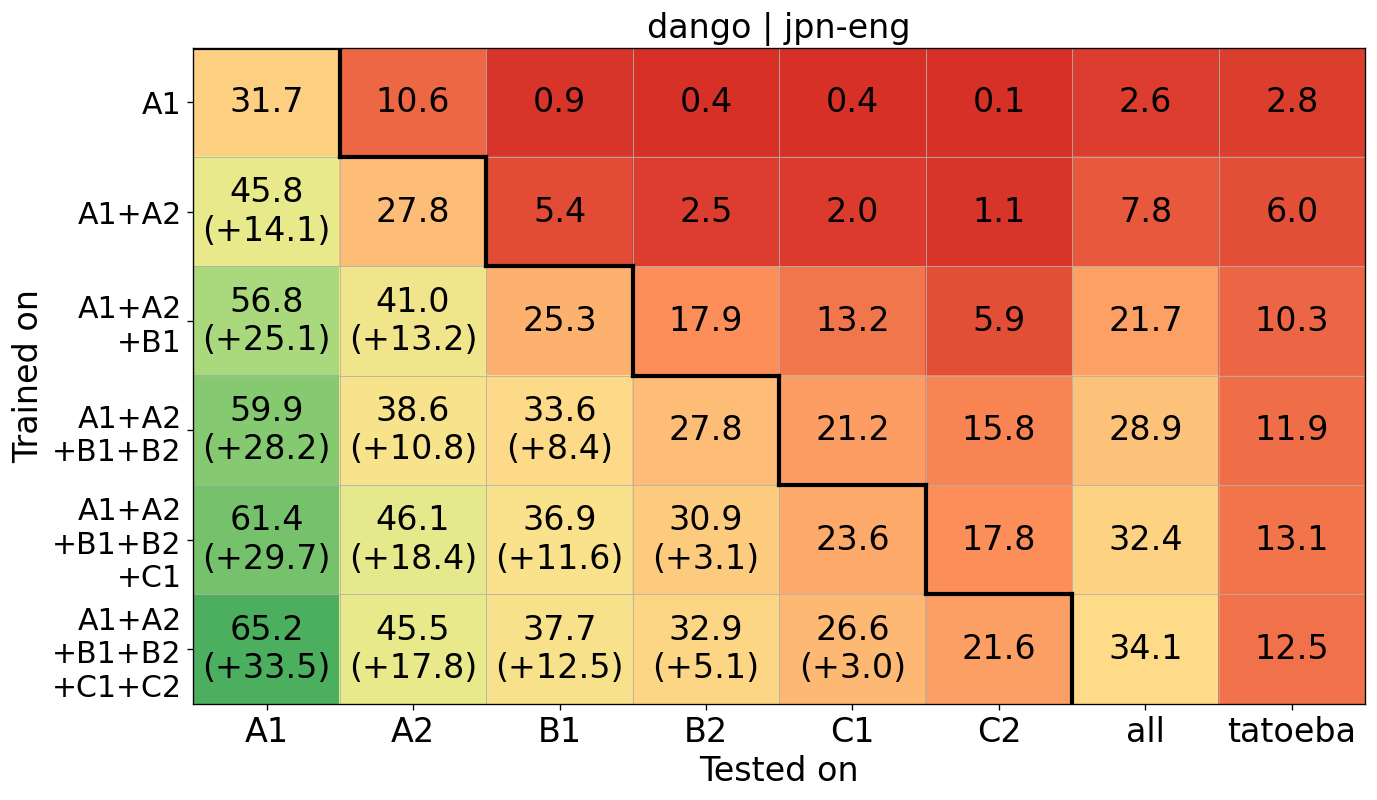}
      \label{fig:learning-jpn-eng-v3}
    \end{subfigure}
    \begin{subfigure}{0.48\textwidth}
      \centering
      \includegraphics[width=\linewidth]{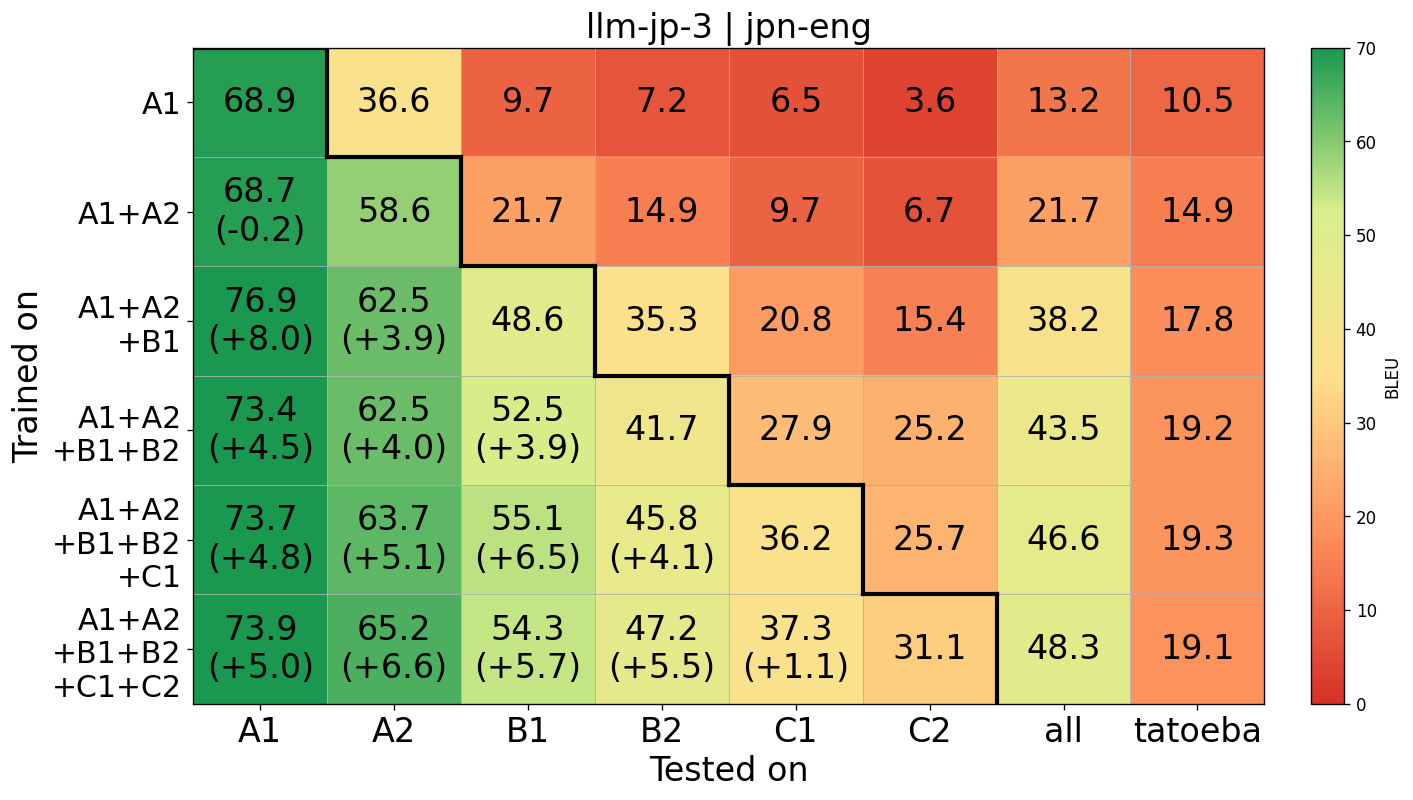}
      \label{fig:learning-jpn-eng-llm-jp-3}
    \end{subfigure}

    \vspace{-6mm}

    \label{fig:learning-jpn-eng}

  \vspace{6pt} 


    
    \vspace{-5mm}

  \caption{Dango and llm-jp-3 Japanese-to-English translation performance when trained and evaluated on different levels of data. The stair-shaped black line separates seen levels (lower left) from unseen levels (upper right).}
  \vspace{-3mm}
  \label{fig:learning-l2}
\end{figure*}

\section{Experiments and Results for L2 (English) Acquisition}

\subsection{Experimental Settings}

\paragraph{Synthesizing L2 Learning Data. }
\label{sec:exp_synth_and_fine-tuning_on_l2}
We used GPT-5.2 to synthesize 9.9K textbook-style L2-learning examples from the CEFR-J word list, generating one example per target word. The dataset comprises 8,938 training, 494 validation, and 501 test examples. See Appendices~\ref{appendix:l2_data_gen_prompt} and~\ref{appendix:l2_data_detail} for prompt design and additional synthesis statistics.

Before fine-tuning on the L2 learning data, we first trained the model on Ichikara, a Japanese instruction-following dataset.\footnote{Visit
\href{https://liat-aip.sakura.ne.jp/wp/llm\%e3\%81\%ae\%e3\%81\%9f\%e3\%82\%81\%e3\%81\%ae\%e6\%97\%a5\%e6\%9c\%ac\%e8\%aa\%9e\%e3\%82\%a4\%e3\%83\%b3\%e3\%82\%b9\%e3\%83\%88\%e3\%83\%a9\%e3\%82\%af\%e3\%82\%b7\%e3\%83\%a7\%e3\%83\%b3\%e3\%83\%87\%e3\%83\%bc\%e3\%82\%bf\%e4\%bd\%9c\%e6\%88\%90/llm\%e3\%81\%ae\%e3\%81\%9f\%e3\%82\%81\%e3\%81\%ae\%e6\%97\%a5\%e6\%9c\%ac\%e8\%aa\%9e\%e3\%82\%a4\%e3\%83\%b3\%e3\%82\%b9\%e3\%83\%88\%e3\%83\%a9\%e3\%82\%af\%e3\%82\%b7\%e3\%83\%a7\%e3\%83\%b3\%e3\%83\%87\%e3\%83\%bc\%e3\%82\%bf-\%e5\%85\%ac\%e9\%96\%8b/}{this link}
for the Ichikara dataset (CC-BY-NC-SA). } This step was intended to strengthen instruction-following behavior in the model’s native language and to enable the case studies in \S~\ref{sec:l1_linguistic_proficiency_assessments} (Appendix~\ref{appendix:case_study_dango_english_after_L1}).

\label{sec:5_fine-tuning_on_L2}
\paragraph{Fine-tuning on L2 Learning Data. }
We trained separate models for six CEFR-J proficiency levels under two direction settings: English--Japanese (Eng--Jpn) and Japanese--English (Jpn--Eng). For a given target level, we used cumulative data from all levels up to and including that level (e.g., B2 uses A1+A2+B1+B2). See Appendix~\ref{appendix:l2_fine-tuning_prompt} and \ref{appendix:l2_sft_detail} for additional fine-tuning details.

\subsection{L2 Proficiency Development Assessment}
Figure~\ref{fig:learning-l2} shows Jpn--Eng translation performance across difficulty levels for Dango and llm-jp-3 trained on different mixtures of difficulties. Results for the Eng--Jpn direction are shown in Appendix~\ref{appendix:l2_eng-jpn}. 
For Tatoeba, we randomly sampled 1,000 sentence pairs from the downloaded dataset.
We provide a human analysis of whether the BLEU score captures meaningful L2 development in Appendix~\ref{appendix:l2_bleu_human_eval}.

\subsubsection{Japanese--English Translation} 
\paragraph{General observations.}
When trained and evaluated on the same difficulty level, for example A1, Dango produces understandable English translations and reaches a BLEU score of 31.7. At this level, llm-jp-3 outperforms Dango with a +37.2 BLEU advantage.

Performance improves as the training and evaluation settings expand. When trained on all levels from A1 to C2 and evaluated on C2, Dango reaches 21.6 BLEU. Qualitatively, it often preserves the core meaning in long sentences, but with frequent lexical and grammatical errors (see Appendix~\ref{appendix:case_study_bleu}). 
Overall, llm-jp-3 consistently outperforms Dango, likely due to its multilingual pretraining.

Both models can generalize to unseen levels and Tatoeba data to some extent. For example, when trained on A1–B2 data, Dango generalizes to C1 (21.2) and C2 (15.8). Being multilingual, llm-jp-3 generalizes better, achieving 27.9 and 25.2.


\paragraph{Training on more difficult data improves performance at easier levels.}
Training with more difficult data consistently enhances performance on easier test sets. For instance, when evaluated on B1, Dango sees a +12.5 BLEU gain when training expands from A1–B1 to A1–C2, whereas llm-jp-3 gains only +5.7 BLEU. 
This demonstrates that Dango learns effectively from the L2-learning data. 

\begin{table}[t]
\centering
\small
\begin{tabular}{llll}
\toprule
\textbf{Model} &
\makecell{\textbf{JSD of}\\\textbf{UF}} &
\makecell{\textbf{JSD of}\\\textbf{ER}} &
\makecell{\textbf{BCD of}\\\textbf{ER}} \\
\midrule

\multicolumn{4}{l}{\textbf{Training-based methods}} \\
\addlinespace[2pt]
Dango
  & 0.0104\bronze{}
  & 0.0110\gold{}
  & 0.2467\silver{} \\
Unfiltered
  & 0.0176
  & 0.0304\bronze{}
  & 0.2502\bronze{} \\
llm-jp-3
  & 0.0249
  & 0.0487
  & 0.6982 \\

\midrule

\multicolumn{4}{l}{\textbf{Prompting-based methods}} \\
\addlinespace[2pt]
llm-jp-3.1-instruct
  & 0.0346
  & 0.0658
  & 0.8523 \\
GPT-4o
  & 0.0028\gold{}
  & 0.0374
  & 0.5295 \\
GPT-5.5
  & 0.0084\silver{}
  & 0.0191\silver{}
  & 0.1925\gold{} \\

\bottomrule
\end{tabular}
\caption{LLM output linguistic feature distribution divergences to human Written Essay produced by Japanese learners. Lower is more similar.}
\label{tab:jsd_human_vs_llm_combined_written_essays}
\end{table}


\begin{table}[t]
\centering
\small
\begin{tabular}{llll}
\toprule
\textbf{Human L1} & \makecell{\textbf{JSD of}\\\textbf{UF}} &
\makecell{\textbf{JSD of}\\\textbf{ER}} &
\makecell{\textbf{BCD of}\\\textbf{ER}}\\
\midrule
Japanese   & 0.0104\gold{} & 0.0110\silver{} & 0.2467\bronze{} \\
\midrule
Korean     & 0.0111\silver{} & 0.0094\gold{} & 0.1078\gold{}\\
Mandarin   & 0.0140\bronze{} & 0.0161\bronze{} & 0.3509 \\
Thai       & 0.0155 & 0.0221 & 0.1272\silver{} \\
\bottomrule
\end{tabular}
\caption{Linguistic feature divergences between Dango and humans with different L1 backgrounds. }
\vspace{-3mm}
\label{tab:l1_diff_written_essay}
\end{table}

\subsection{Transfer Effect Assessment}
\label{sec:human_likeness_assessment}

\paragraph{Baselines.}
In addition to the training-based methods (Dango, unfiltered, and llm-jp-3), we adopt the prompting-based L2 speaker simulation method from \citet{gao-etal-2025-llms}, where an LLM is instructed to generate L2 English text conditioned on prompts containing a specified L1 background and L1-dependent linguistic traits. We evaluate three LLMs in this setting: llm-jp-3.1-instruct,\footnote{We used the 1.8B model: \url{https://huggingface.co/llm-jp/llm-jp-3.1-1.8b-instruct4}} GPT-4o,\footnote{\url{https://openai.com/index/hello-gpt-4o/}} and GPT-5.5.\footnote{\url{https://openai.com/index/introducing-gpt-5-5/}} For GPT-4o and GPT-5.5, we follow the original prompt design, modifying it to output English translations (Figure~\ref{fig:llm_l2_roleplay_user_prompt} and~\ref{fig:llm_l2_roleplay_system_prompt}); for llm-jp-3.1-instruct, we use a simplified role-playing and error-injection prompt (Figure~\ref{fig:llm-jp-3.1-instruct_l2_roleplay_user_prompt}) due to its smaller size and shorter context limit. 

Both approaches offer distinct advantages. Prompting-based methods are highly accessible, utilizing off-the-shelf LLMs without the need for additional training. Conversely, training-based models allow us to trace the transfer effect directly back to the L1→L2 training process itself. This isolates the learning mechanism from the confounding effects of an LLM simply ``role-playing'' to mask its actual proficiency, providing a stronger foundation for future controlled studies and interpretability research.

As described in \S~\ref{sec:3_transfer_effect_assessment}, we first construct UF and ER distributions for human and model English productions. We then compute JSD and BCD for human--model pairs to identify which models best match human transfer patterns. Here, we report results on ICNALE Written Essays; Appendix~\ref{appendix:l2_spoken_dialogues} presents the corresponding analysis on ICNALE Spoken Dialogues, confirming the robustness of our findings. For trained baselines, we use models trained on A1--B2 data, since the ICNALE learners are below B2 level.

\paragraph{Focusing on JSD of UF. } Table~\ref{tab:jsd_human_vs_llm_combined_written_essays} shows that, among trained baselines, Dango is closer to Japanese L1 speakers than the unfiltered model or llm-jp-3 in terms of UF divergence. In other words, Dango has more human-like overuse and avoidance patterns over the seven feature categories. 

Compared with prompting baselines, GPT-4o and GPT-5.5 show UF distributions closer to humans, slightly outperforming Dango. In contrast, llm-jp-3.1-instruct remains the most distant, likely due to its small size and limited exposure to such role-play prompts during training. It generally failed to output systematic English grammar errors, even when prompted to do so. 

Table~\ref{tab:l1_diff_written_essay} shows Dango’s UF divergence from human English learners across L1 backgrounds. The pattern broadly aligns with the degree of structural overlap with Japanese (Table~\ref{tab:lang_features}): Dango is most similar to Japanese learners, followed by Korean, likely due to shared SOV word order and similar particle systems. Mandarin shows greater divergence, while Thai diverges the most, reflecting its limited structural similarity to Japanese.


\begin{figure}[t]
    \centering

    \includegraphics[width=0.99\linewidth]{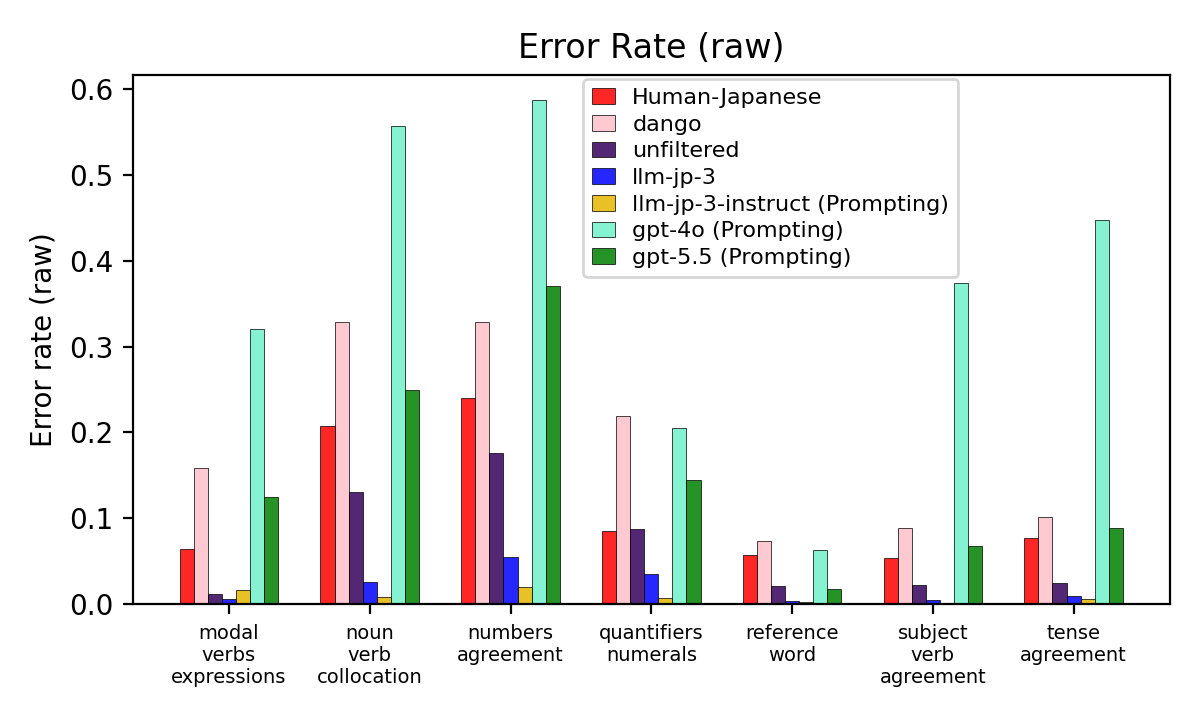}

    \vspace{-2mm}
    \caption{Error Rates (raw values) of human and LLMs.}
    \vspace{-2mm}
    \label{fig:error_dist}
\end{figure}

\paragraph{Focusing on JSD of ER. } In Table~\ref{tab:jsd_human_vs_llm_combined_written_essays}, Dango achieves the lowest ER JSD among all methods, indicating that its error distribution is closest to that of human learners. Compared with non-Japanese L1 learners (Table~\ref{tab:l1_diff_written_essay}), Dango is closest to Korean speakers, followed by Japanese speakers. We attribute this to the close linguistic distance between Japanese and Korean, and interpret it as evidence that Dango captures L1-induced transfer effects.



\paragraph{Focusing on BCD of ER. } Dango achieves the second-lowest ER BCD in Table~\ref{tab:jsd_human_vs_llm_combined_written_essays}. GPT-5.5 obtains the lowest BCD and remains balanced across the other metrics, suggesting that it also models Japanese--English transfer effects well. In contrast, the other multilingual baselines show large BCD values because their absolute error rates differ substantially from human learners (Figure~\ref{fig:error_dist}): llm-jp-3 and llm-jp-3.1-instruct rarely make grammar errors, whereas GPT-4o produces too many.

Overall, these results suggest that Dango’s English productions broadly align with a Japanese L1 speaker profile. Together with the additional analysis in Appendix~\ref{appendix:l2_spoken_dialogues}, they show that filtering L2 exposure in the Japanese corpus improves L2 speaker modeling, as the unfiltered baseline performs worse than Dango on all measured metrics.

\subsection{Analysis on Dango's English Production}
We present a case study on Dango's English production in Figure~\ref{fig:case_study_dango_l2_errors}, with a focus on its error patterns. We observe that Dango makes many grammatical and phrasing errors typical of Japanese speakers, such as preposition, article form, and tense errors. We also observed that it sometimes mistranslates Japanese words literally into English, mirroring a strategy L2 speakers often use when they lack the vocabulary for a specific concept \cite{faerch1983strategies}.

\section{Conclusions}
We presented \textbf{Dango}, a 1.8B-parameter Japanese L1-only decoder LLM trained on a strictly filtered corpus to study L1→L2 transfer. By fine-tuning on LLM-generated English-learning lessons, Dango exhibits learner-like transfer patterns closer to Japanese L2 speakers than other trained baselines, on par with prompted GPT-5.5. We release Dango to advance controlled SLA studies and facilitate practical learner-facing applications.


\section*{Limitations}
\paragraph{Tokenizer and vocabulary.}
Dango uses a pretrained multilingual tokenizer with substantial English subword coverage. This choice improves training stability and enables direct comparison to llm-jp-3, but it also means Dango does not fully learn its vocabulary from scratch under an L1-only constraint. Importantly, however, the token embeddings themselves are learned from scratch during pretraining. Future work could explore tokenizer choices and dynamic vocabulary expansion schemes during L2 exposure to better approximate staged, human-like acquisition.

\paragraph{Single model and single L1→L2 setting.}
We study a single model scale (1.8B parameters) due to the high cost of pretraining, and we focus on one transfer direction (Japanese$\rightarrow$English) because English is a common and practically important L2. Future work can extend this framework to larger model scales and additional L1$\rightarrow$L2 pairs spanning a wider range of linguistic distances to better characterize the generality of our findings.

\section*{AI Assistance in This Work}
We utilized AI-based tools primarily to refine the writing style and improve the grammatical accuracy of this manuscript. Additionally, we employed AI assistance for data visualization and minor coding support. The authors carefully reviewed and verified all AI-assisted content, figures, and code, and we take full responsibility for the final manuscript and its outcomes.




\begin{CJK*}{UTF8}{min}
\bibliography{custom}
\end{CJK*}

\clearpage

\appendix

\section{Filtering Pipeline Design and Statistics}
\label{appendix:filtering_details}
Our filtering pipeline for constructing a high-quality Japanese pretraining corpus consists of three stages: (1) a Unicode whitelist filter, (2) an English (Latin-script) blocklist filter, and (3) a Chinese blocklist filter. Table~\ref{tab:filter_params} summarizes the parameter settings used in each stage.

\paragraph{Unicode whitelist filter. } This filter drops a document if the fraction of characters outside a predefined Unicode whitelist exceeds a threshold.
The whitelist includes Japanese kana, CJK ideographs, punctuation, emojis, and ASCII characters (which may include Latin letters). CJK ideographs contains Chinese characters used by Chinese, Japanese (as Kanji), and Korean (as Hanja). Therefore, Korean Hanguls does not pass the whitelist filter. 
Chinese characters and English (Latin) characters may pass this stage, and are handled by the subsequent blocklist filters.
For other languages, if a document contains, e.g., Cyrillic, characters totaling more than 0.1\% of all characters, the entire document is removed from the corpus.

\paragraph{English (Latin) filter. }
This filter operates at the line level (lines are split by ``\textbackslash n'').
It removes lines that contain excessive Latin-script content, and drops the entire document if too many lines are removed (to avoid broken context).
A line is removed if any of the following conditions hold:
(i) it contains more than 20 Latin characters;
(ii) the ratio of Latin to non-Latin characters exceeds 40\%;
or (iii) it contains more than four consecutive Latin words separated by spaces.
Lines containing fewer than eight Latin characters are ignored, as we consider such cases harmless.
Finally, if more than 5\% of lines in a document are removed, we drop the document.

\paragraph{Chinese filter. }
To reduce Chinese text (primarily Simplified Chinese) in the corpus, we construct a character-level blocklist of ideographs that are highly indicative of Chinese-only usage.
The blocklist is derived from the Unihan BMP list.\footnote{\texttt{https://unicode.org/charts/unihan.html}}
If any character from this blocklist appears in a line, the line is removed.
If more than 0.1\% of lines are removed from a document, the document is dropped.

\paragraph{Filtering outcome.}
Applying the full pipeline removes approximately 30\% of documents from the original corpus, highlighting the need for strict filtering when building a Japanese L1 pretraining dataset. After filtering, the corpus's total token count shrank from 592B to 376B.

\section{Filtered Documents: Case Study}
\label{appendix:filtering_cases}
The English (Latin-script) blocklist filter rejects most cases of severe English contamination in the Japanese corpus, including document-level parallel data (Figure~\ref{fig:contamination_document_level_parallel}) and phrase-level parallel data (Figure~\ref{fig:contamination_word_level_parallel}). In these examples, lines (separated by ``\textbackslash n'') shown in red are flagged by the filter. Because the fraction of flagged lines exceeds 5\% in each document, the entire document is then removed. 

At the same time, the pipeline permits limited English that a typical Japanese speaker may encounter or use in everyday life. For instance, in Figure~\ref{fig:allowed_english}, common short tokens such as ``hit'' and ``ok,'' as well as proper nouns such as ``Access'' and ``Excel,'' do not trigger the blocklist. Short phrases (e.g., ``stay at home order,'' ``Reasons I Love You'') are also allowed; these represent an upper bound on the length of consecutive English segments that Dango is likely to have encountered during training.

\newcommand{\shown}{\textbf{\textbackslash n}}

\begin{figure}
    \centering
\begin{tcolorbox}[width=\linewidth, colback=white, colframe=black!30, boxrule=0.4pt]
\begin{CJK}{UTF8}{min} 
\scriptsize\ttfamily
\raggedright
\parindent=0pt
\begingroup
\let\n\shown
\obeylines
(......)
若年女性の流行
Trends
{\color{red}
Japan is the land of trends. Nowhere else do trends arise, spread and \n
die with such speed. The reasons for this are simple: affluent youth, merciless \n
advertising, high population density and an insatiable appetite for 'the \n
}
(......)
流行
日本は流行り廃りの激しい国だ。こんなに速いスピードで流行が生まれ、広がり、そして廃れていく国は他にない。そうなった理由は単純だ。小金を持った若年層、容赦ない宣伝、高い人口密度、そして新し物好きの国民性。  \n
(......)
\endgroup
\end{CJK}
\end{tcolorbox}
    \caption{Contamination: document-level parallel data.}
    \label{fig:contamination_document_level_parallel}
\end{figure}

\begin{figure}
    \centering
\begin{tcolorbox}[width=\linewidth, colback=white, colframe=black!30, boxrule=0.4pt]
\begin{CJK}{UTF8}{min} 
\scriptsize\ttfamily
\raggedright
\parindent=0pt
\begingroup
\let\n\shown
\obeylines
(......)
左ひだり left (side) \n
{\color{red}折おれて turn (toward) \n}
{\color{red}血塔けっとう Bloody Tower \n}
門もん gate \n
入はいる enter \n
{\color{red}今いま now; the present \n}
{\color{red}昔むかし long ago; ancient times \n}
{\color{red}薔薇しょうびの乱らん War of the Roses (15th century struggle for the throne of England) \n}
(......)
\endgroup
\end{CJK}
\end{tcolorbox}
    \caption{Contamination: word/phrase-level parallel data.}
    \label{fig:contamination_word_level_parallel}
\end{figure}

\begin{figure}
    \centering
\begin{tcolorbox}[width=\linewidth, colback=white, colframe=black!30, boxrule=0.4pt]
\begin{CJK}{UTF8}{min} 
\scriptsize\ttfamily
\raggedright
\parindent=0pt
\begingroup
\let\n\shown
\obeylines
アクセス数:0 hit (累計:15524 hit) 逆アクセス数:0 hit (累計:6371 hit)
------------------------------------
応募資格 IT関連実務経験が少しでもあればok！Access、Excelマクロ経験ある方は尚歓迎です！／　◆ブランク3年程度でもご相談ください◆
------------------------------------
ハワイでは、3月22日に州政府より緊急事態宣言が発令され、翌日午後4時より4月30日まで不要不急の外出を禁止する “stay at home order”が発令されました。この法令は、生活に不可欠なサービス・商品を提供している機関を除き、原則すべての従業員は在宅勤務を行わなければならない、というものでした。
------------------------------------
今でも年に2回は海外へ行っているという西野カナさん。2016年の新曲「あなたの好きなところ」は、52枚のトランプに好きな人の好きなところを書いてプレゼントするラブレター"52 Reasons I Love You"　に着想を得て生まれた曲だそう。
\endgroup
\end{CJK}
\end{tcolorbox}
    \caption{English appearances that are allowed by the filtering pipeline.}
    \label{fig:allowed_english}
\end{figure}





\section{L1 Model Pretraining Details}
\label{appendix:pre-training}
We reproduced the customized Megatron-LM pretraining framework developed by llm-jp.\footnote{\url{https://github.com/llm-jp/scripts/tree/main/pretrain/scripts}} Most hyperparameters, such as model architecture and optimization settings, were kept identical to those of the official llm-jp-3 models (Table~\ref{tab:training_details}), except that we increased the global batch size from 512 to 1024. We set the learning-rate schedule to a maximum of 376B tokens to match the size of the L1-stage Japanese corpus, but trained for only 100B tokens. In contrast, the final released llm-jp-3 model is scheduled for 2.1T tokens in total. Thus, our model and the corresponding intermediate llm-jp-3 checkpoint represent partial progress along their respective full training trajectories. Pretraining was conducted on 32 NVIDIA H200 GPUs (4 nodes × 8 GPUs) for approximately 26 hours.

\begin{table}[h]
    \centering
    \small
    \begin{tabular}{lr}
        \toprule
        \textbf{Hyperparameter} & \textbf{Value} \\
        \midrule
        \multicolumn{2}{l}{\textit{Model Architecture}} \\
        \quad Layers & 24 \\
        \quad Hidden Size & 2048 \\
        \quad FFN Hidden Size & 7168 \\
        \quad Attention Heads & 16 \\
        \quad Sequence Length & 4096 \\
        \quad Positional Embeddings & RoPE \\
        \quad Activation Function & SwiGLU \\
        \quad Normalization & RMSNorm \\
        & ($\epsilon=1 \times 10^{-5}$) \\
        \quad Bias & Disabled \\
        \midrule
        \multicolumn{2}{l}{\textit{Optimization}} \\
        \quad Optimizer & Adam \\
        & ($\beta_1, \beta_2=0.9, 0.95$) \\
        \quad Peak Learning Rate & $3 \times 10^{-4}$ \\
        \quad Min Learning Rate & $3 \times 10^{-5}$ \\
        \quad LR Scheduler & Cosine \\
        \quad Warmup Steps & 180 \\
        \quad Weight Decay & 0.1 \\
        \quad Gradient Clipping & 1.0 \\
        \quad Global Batch Size & 1024 \\
        \quad Precision & bf16 \\
        \quad Flash-attention & True \\
        \quad Z-Loss Strength & $1 \times 10^{-4}$ \\
        \bottomrule
    \end{tabular}
    \caption{Model architecture and training hyperparameters.}
    \label{tab:training_details}
\end{table}

\section{Category Breakdown of Tasks in llm-jp-eval}
\label{appendix:llm-jp-eval-radar}
We exclude Code Generation (CG) and Mathematical Reasoning (MR) from the evaluation suite because they are not directly related to our goal. We also exclude Machine Translation (MT; Japanese–English) because we intend to focus on evaluating Japanese proficiency using llm-jp-eval; we leave English evaluation to MultiBLiMP.

\begin{figure}[t]
    \centering

    \includegraphics[width=0.95\linewidth]{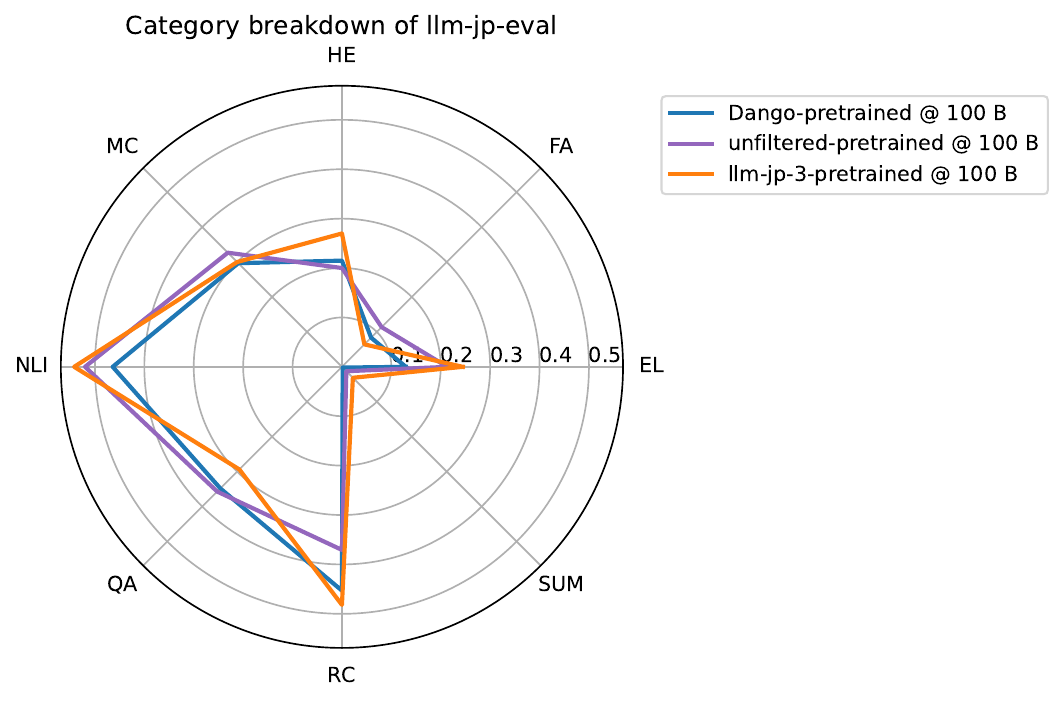}

    \vspace{-2mm}
    \caption{Task category breakdown of llm-jp-eval performance. }
    \label{fig:l1_llm-jp-eval_radar}
\end{figure}

In Figure~\ref{fig:l1_llm-jp-eval_radar}, we find that models show mixed performance across task categories, with each model outperforming the other in different categories. 

\section{Case Study on Dango's English Competency After L1 Acquisition}
\label{appendix:case_study_dango_english_after_L1}
\begin{CJK}{UTF8}{min}
After completing L1 pretraining, we fine-tuned Dango-pretrained on the Ichikara instruction-following dataset, which includes a small number of Japanese--English translation examples. We then manually inspected the models’ English generation (translation) behavior.

As shown in the upper block of Figure~\ref{fig:case_study_n_gram_level}, for a relatively simple Japanese--English translation prompt, Dango occasionally captures local lexical correspondences (e.g., mapping ``犬'' to ``dog''), but its output is typically unfaithful and ungrammatical at the sentence level. In contrast, llm-jp-3 captures more nuances and the grammar is correct. 

For a longer sentence (bottom block), Dango fails to preserve the meaning of the Japanese source and instead generates largely unrelated and repetitive English text. Nevertheless, parts of the output contain locally well-formed fragments (e.g., ``you are not'', ``I love you''), suggesting that the four-consecutive-English-word filter encourages learning short-range English regularities (n-gram-level patterns) rather than maintaining global sentence semantics. llm-jp-3 also fails to translate the sentence faithfully in this case, but its output remains grammatically coherent and stylistically natural.
\end{CJK}

\newcommand{\segline}{%
  \par\smallskip
  \noindent\textcolor{black!30}{\rule{\linewidth}{0.4pt}}%
  \par\smallskip
}

\begin{figure}
    \centering
\begin{tcolorbox}[width=\linewidth, colback=white, colframe=black!30, boxrule=0.4pt]
\begin{CJK}{UTF8}{min} 
\scriptsize\ttfamily
\raggedright
\parindent=0pt
\begingroup
\let\n\shown
\obeylines

\textbf{User}: 
「私は可愛い犬がいるカフェに行きました。」を英語で書いてください。
{\color{gray}{(translation: Write "I went to a cafe where there were cute dogs." in English. )}}

\segline

\textbf{Dango-pretrained-Ichikara}: 
「私は可愛い犬がいるカフェに行きました。」を英語に翻訳すると、以下のようになります。
I am beautiful dog. 
{\color{gray}{(translation: If you translate "......" to English, it becomes \textbackslash n I am beautiful dog. )}}

\segline

\textbf{llm-jp-3-pretrained-Ichikara}: 
I'm looking for a cute dog restaurant.

\endgroup
\end{CJK}
\end{tcolorbox}

\begin{tcolorbox}[width=\linewidth, colback=white, colframe=black!30, boxrule=0.4pt]
\begin{CJK}{UTF8}{min} 
\scriptsize\ttfamily
\raggedright
\parindent=0pt
\begingroup
\let\n\shown
\obeylines
\textbf{User}: 
「彼女は、プレッシャーの中でも落ち着いていられる彼の能力を感心して話した。」を英語で書いてください。
{\color{gray}{(translation: Write "She spoke admiringly of his ability to stay calm under pressure." in English. )}}

\segline

\textbf{Dango-pretrained-Ichikara}:
""He is a hot day to meet you are not happen again’s hot day to meet you are not happen again’s hot day to meet you are not been happen. I love you, I love you."

のように書くと良いでしょう。
{\color{gray}{(translation: You can write it as "......"  )}}
\segline

\textbf{llm-jp-3-pretrained-Ichikara}:
「彼女は、プレッシャーの中でも落ち着いていられる彼の能力を感心して話した。」を英語に翻訳すると、以下のようになります。

I enjoyed listening to the story of the day because the conflict was resolved in a healthy way.
{\color{gray}{(translation: If you translate "......" to English, it becomes \textbackslash n ...... )}}

\endgroup
\end{CJK}
\end{tcolorbox}
    \caption{Case study of English production after completing L1 pretraining and Ichikara instruction-tuning. }
    \label{fig:case_study_n_gram_level}
\end{figure}

\section{L2 Acquisition Data Synthesis Details}

\subsection{Prompt for L2 Learning Data Generation}
\label{appendix:l2_data_gen_prompt}
\begin{figure*}[t]
  \centering
  \includegraphics[width=0.97\textwidth]{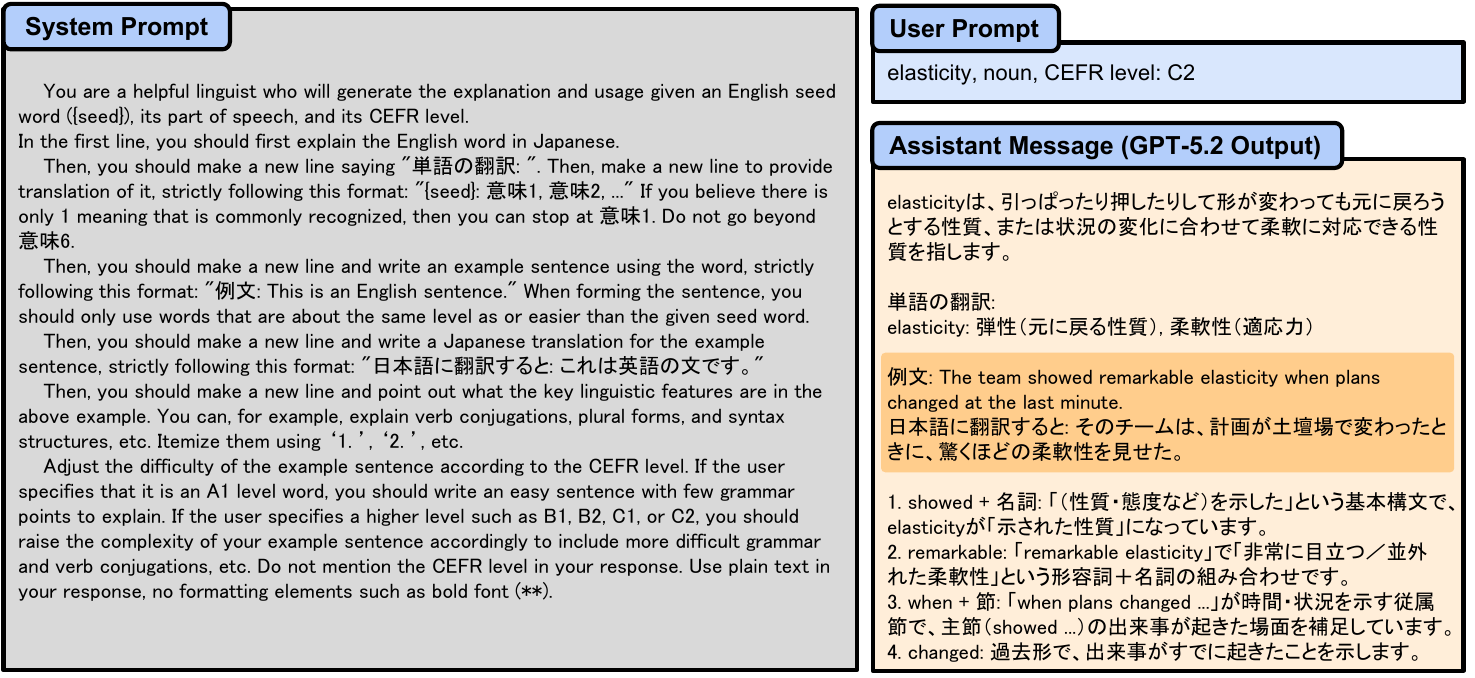}
  \vspace{-3mm}
  \caption{Prompt used to generate textbook-style L2 learning data. }
  \label{fig:l2_data_gen_prompt}
  \vspace{-5mm}
\end{figure*}

Figure~\ref{fig:l2_data_gen_prompt} showcases the prompt used to generate  textbook-style L2 learning data with GPT-5.2. The deep orange block highlights the parallel sentence pair, which is used in evaluation (Figure~\ref{fig:l2_test_prompt}).

\subsection{Prompt for L2 Acquisition Fine-tuning}
\label{appendix:l2_fine-tuning_prompt}
We show the prompt format used to fine-tune the models on the L2 learning data. We modified the user prompt and fine-tuned the models to learn GPT-5.2’s outputs; the assistant message is identical to that in Figure~\ref{fig:l2_data_gen_prompt}.

\begin{figure}[t]
  \centering
  \includegraphics[width=0.90\linewidth]{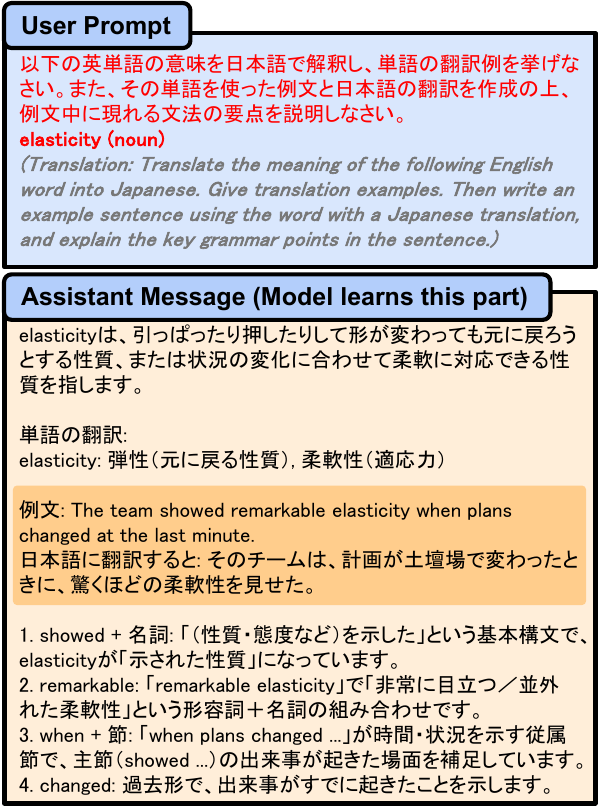}
  \vspace{-3mm}
  \caption{Prompt format used to train Dango and llm-jp-3 on the textbook-style L2 learning data. }
  \label{fig:l2_fine-tuning_prompt}
  \vspace{-5mm}
\end{figure}

\subsection{L2 Learning Data Length Statistics}
\label{appendix:l2_data_detail}
\begin{figure}[t]
    \centering

    \includegraphics[width=0.90\linewidth]{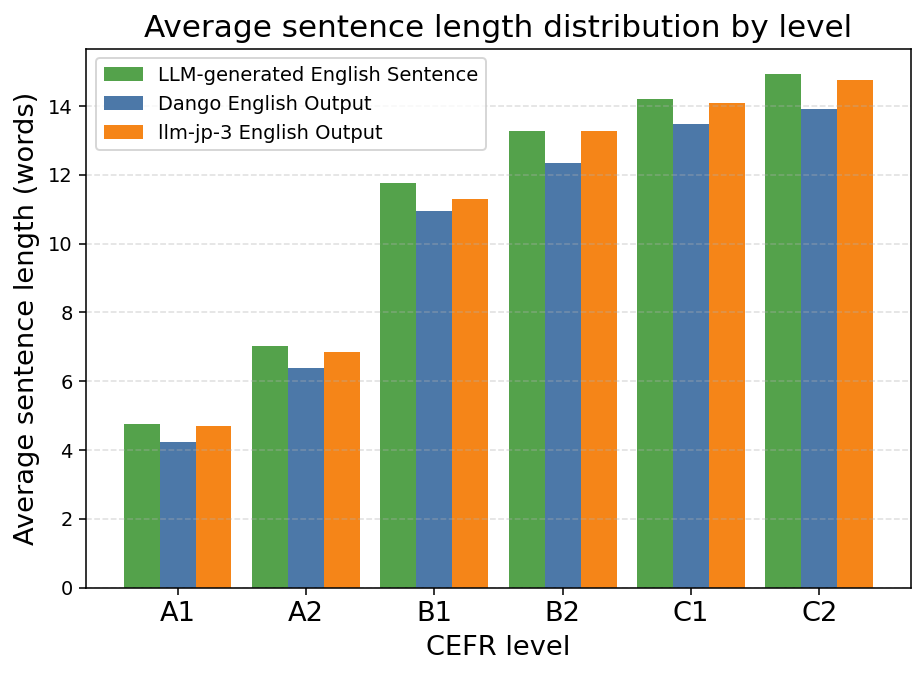}

    \vspace{-2mm}
    \caption{Sentence length distribution of the LLM-generated L2 learning data. }
    \label{fig:l2_learning_data_sentence_len_dist}
\end{figure}

In Figure~\ref{fig:l2_learning_data_sentence_len_dist}, we report the average sentence length (in words) in the test set of our LLM-generated L2 learning data. For comparison, we also report the sentence lengths of Dango and llm-jp-3’s English outputs from the Japanese--English translation task. For each CEFR-J level (e.g., B2), outputs are sampled from models trained on data up to that level (e.g., A1+A2+B1+B2).

\subsection{Prompt for Evaluating Models on Translation Task}
\label{appendix:l2_translation_eval_prompt}
\begin{figure}[t]
  \centering
  \includegraphics[width=0.90\linewidth]{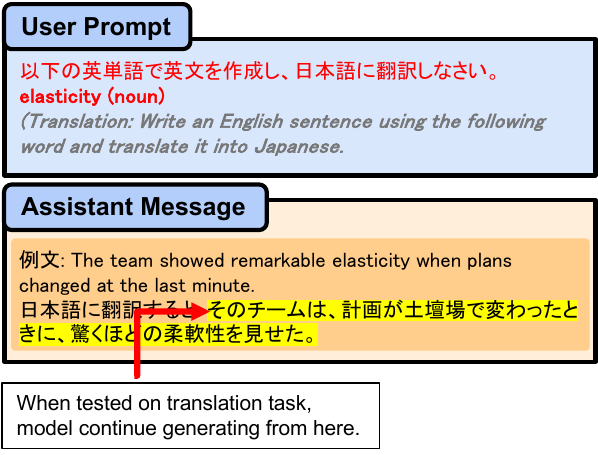}
  \vspace{-3mm}
  \caption{Prompt format used to evaluate the models on the translation task. The model continues generating after the source sentence appears in the assistant message, only producing the highlighted yellow part. }
  \label{fig:l2_test_prompt}
\end{figure}

We show the prompt format used to evaluate the translation task in Figure~\ref{fig:l2_test_prompt}. We place the source sentence in the assistant message rather than in the user prompt, and the model continues generating after “Translate this into Japanese/English:” (highlighted in yellow). We adopt this format because it more closely matches what models see during L2 acquisition: the example sentence and its translation appear together in the assistant message rather than in the user prompt. Empirically, we find that this format yields more stable translation outputs.

\section{L2 Acquisition Fine-tuning Details}
\label{appendix:l2_sft_detail}
We fine-tuned the LLMs using Axolotl.\footnote{\url{https://github.com/axolotl-ai-cloud/axolotl/}} All experiments were conducted on a single GPU without early stopping. The fine-tuning hyperparameters are shown in Table~\ref{tab:fine-tuning_hyperparams}.

\begin{table}[h]
    \centering
    \small
    \begin{tabular}{lr}
        \toprule
        \textbf{Hyperparameter} & \textbf{Value} \\
        \midrule

        \multicolumn{2}{l}{\textit{Training Setup}} \\
        \quad Sequence Length & 4096 \\
        \quad Sample Packing & False \\
        \quad Num Epochs & 3 \\
        \quad Micro Batch Size & 4 \\
        \quad Gradient Accumulation Steps & 4 \\

        \midrule
        \multicolumn{2}{l}{\textit{Optimization}} \\
        \quad Optimizer & AdamW \\
        \quad Learning Rate & $2 \times 10^{-4}$ \\
        \quad LR Scheduler & Cosine \\
        \quad Warmup Ratio & 0.1 \\
        \quad Weight Decay & 0.1 \\

        \bottomrule
    \end{tabular}
    \caption{Fine-tuning hyperparameters.}
    \label{tab:fine-tuning_hyperparams}
\end{table}

\section{L2 Development Assessment: English--Japanese Translation}
\label{appendix:l2_eng-jpn}
\begin{figure*}[!htp]
  \centering

    \centering
    \begin{subfigure}{0.468\textwidth}
      \centering
      \includegraphics[width=\linewidth]{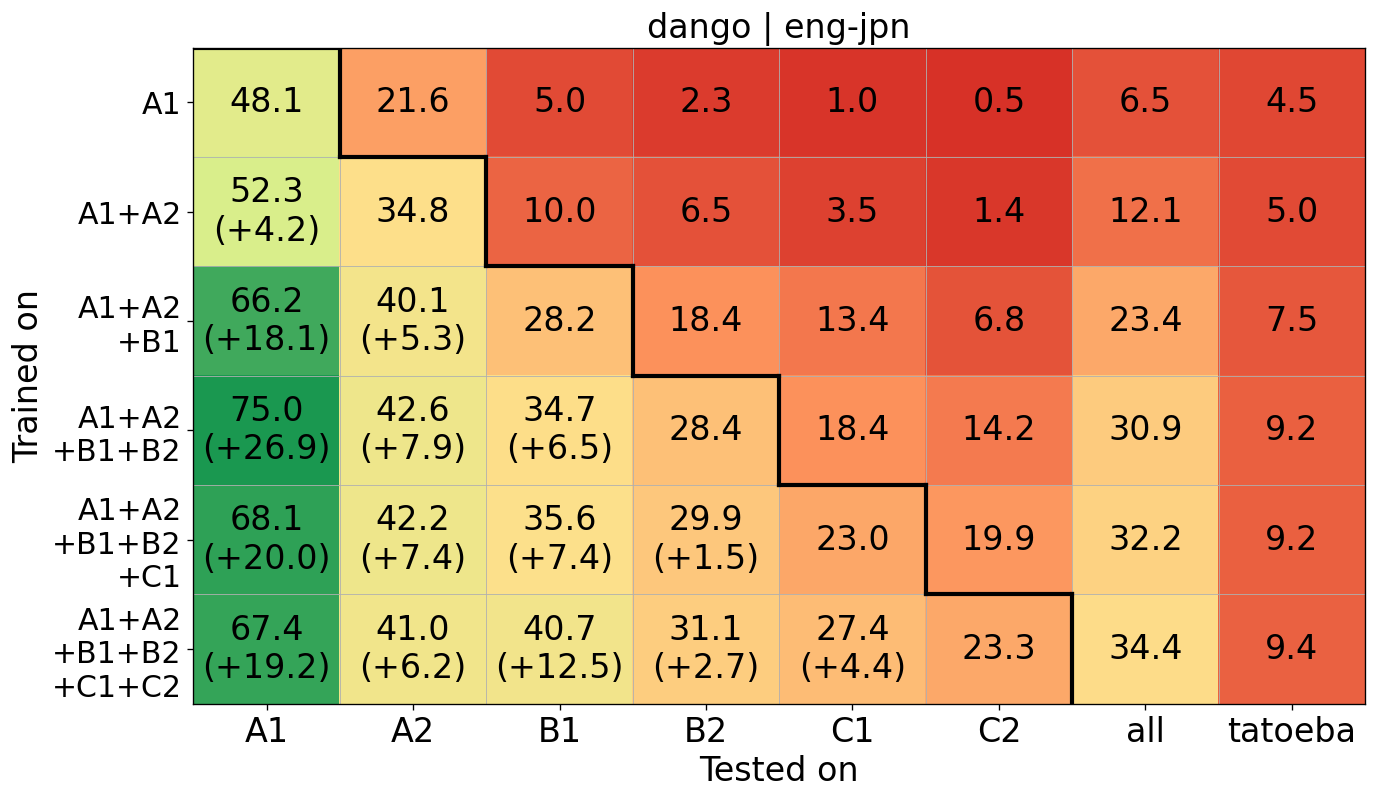}
      \label{fig:learning-eng-jpn-v3}
    \end{subfigure}
    \begin{subfigure}{0.48\textwidth}
      \centering
      \includegraphics[width=\linewidth]{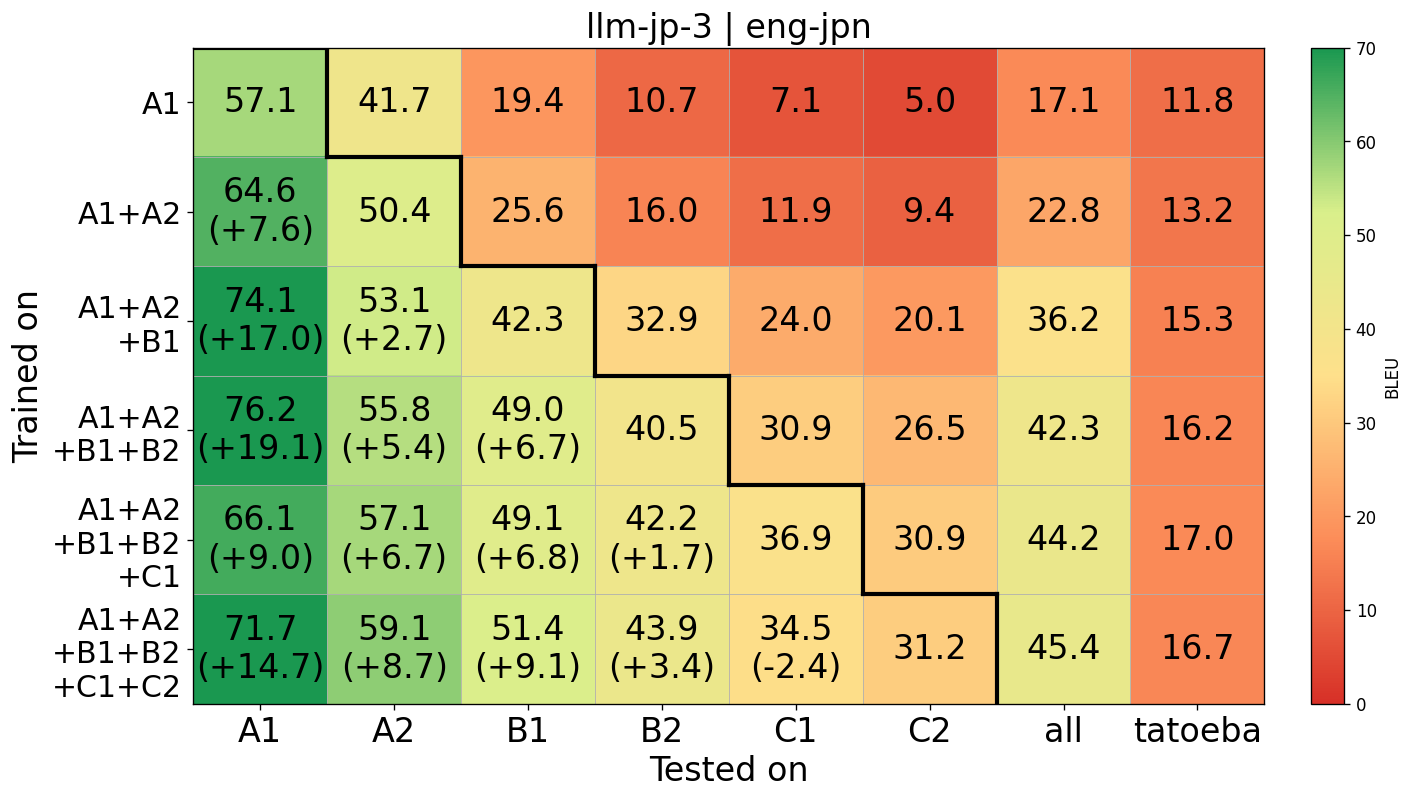}
      \label{fig:learning-eng-jpn-llm-jp-3}

    
    \label{fig:learning-eng-jpn}
  \end{subfigure}
    \vspace{-5mm}

  \caption{Dango and llm-jp-3 when trained and evaluated on different levels of Japanese-English data, showing L2 development progress. The stair-shaped black line separates seen levels (lower left) from unseen levels (upper right).}
  \label{fig:learning-l2-en-ja-appendix}
\end{figure*}

We observe similar trends in the Eng--Jpn direction in Figure~\ref{fig:learning-l2-en-ja-appendix}. Because Dango is native in Japanese, it reaches strong performance on easier levels more quickly than in the Jpn--Eng direction. BLEU saturates earlier, and the performance gap relative to llm-jp-3 is smaller than in Japanese--English translation. However, Dango still struggles with generalization to unseen difficulty levels compared to llm-jp-3, especially when evaluated on harder data than it has seen during training. 

\section{L2 Development Assessment: Does BLEU Capture Proficiency Development?}
\label{appendix:l2_bleu_human_eval}
We conduct a human evaluation to examine whether the observed BLEU improvement reflects meaningful gains in linguistic development, rather than only changes in an automatic metric. 

To assess the validity of BLEU as a proxy, we perform a blinded human evaluation across two Dango checkpoints trained on different proficiency ranges. Specifically, we compare a checkpoint trained on A1--B1 data with one trained on A1--B2 data, and evaluate both on B2-level test examples. This setting corresponds to ``Dango, trained on A1+A2+B1 / A1+A2+B1+B2, tested on B2'' in Figure~\ref{fig:learning-l2}. Since B2 is out-of-distribution for the A1--B1 checkpoint, this comparison probes the seen/unseen proficiency boundary and provides an informative test of whether additional proficiency-level training improves adequacy.

A human annotator (one of the author) with C2-level English proficiency, who is a non-native speaker, blindly (not knowing which model produced the output) rated each model output on a 1--5 correctness scale, with the GPT-5.2 generated translation as gold reference. A score of 5 indicates a fully correct translation with no omissions or unrelated content, 3 indicates that at least one key meaning unit is missing or mistranslated, and 1 indicates that the output is completely incorrect or unrelated. 

\begin{table}[t]
\centering
\small
\begin{tabular}{lcc}
\toprule
\textbf{Model} & \textbf{BLEU} & \textbf{Avg. human correctness} \\
& & \textbf{(1--5)} \\
\midrule
Dango [A1--B1] & 17.9 & 2.76 \\
Dango [A1--B2] & 27.8 & 3.04 \\
\bottomrule
\end{tabular}
\caption{Human evaluation results on the B2 test set ($n=50$).}
\label{tab:human-eval-b2}
\end{table}

As shown in Table~\ref{tab:human-eval-b2}, the checkpoint with the higher BLEU score also receives a higher average human correctness score. This suggests that, in this setting, BLEU tracks meaningful improvements in translation adequacy, although it should still be interpreted as an automatic proxy rather than a direct measure of L2 proficiency development.

\subsection{Case Study on Error Pattern and Proficiency Development}
\label{appendix:case_study_bleu}
We present a case study examining Dango’s error patterns and proficiency development. As shown in Figure~\ref{fig:case_study_cefr_english_quality}, Dango exhibits human-like L2 errors, such as article omission and non-native phrasing. As it is exposed to more training data, it successfully learns to avoid these mistakes.

\section{L2 Human-like Linguistic Feature Assessment Details}
We adopted the framework of \citet{gao-etal-2025-llms} to extract linguistic features and measure their usage frequency and error rate in model translations and human English productions. Following their setup, we used GPT-4o as the annotator with temperature set to 0, and parsed each instance once.

\paragraph{Prompt Designs.} Since \citet{gao-etal-2025-llms} originally designed prompts for dialogue annotation, we adapted their prompts for single-sentence model outputs by instructing the LLM to annotate one sentence at a time.

For human English productions, we further adapted the prompts for written essays in the main experiment. Each essay was split into 5-sentence chunks, and sentences with fewer than 4 words were skipped. For the extended analysis on ICNALE Spoken Dialogues (Appendix~\ref{appendix:l2_spoken_dialogues}), we used the original dialogue-based prompt design. The full prompt set follows \citet{gao-etal-2025-llms} and their repository.\footnote{\url{https://github.com/RenaGao/LLMPirorknowledge/tree/main}}

\begin{figure*}[t]
  \centering
  \includegraphics[width=0.95\textwidth]{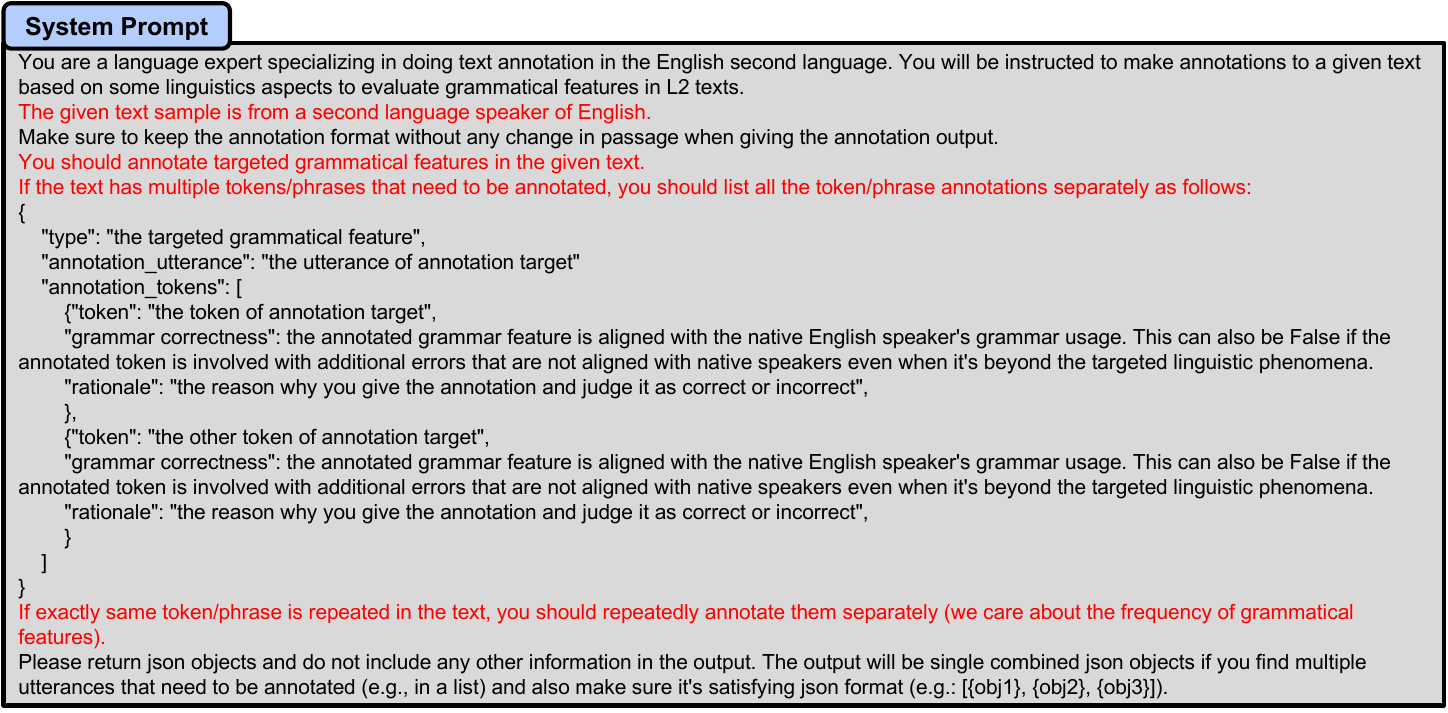}
  \vspace{-3mm}
  \caption{System prompt for linguistic feature annotation. We build on the framework of \citet{gao-etal-2025-llms}; text in red indicates our modifications.}
  \label{fig:annotation_system_prompt}
  \vspace{-5mm}
\end{figure*}

\begin{figure}[t]
  \centering
  \includegraphics[width=0.97\linewidth]{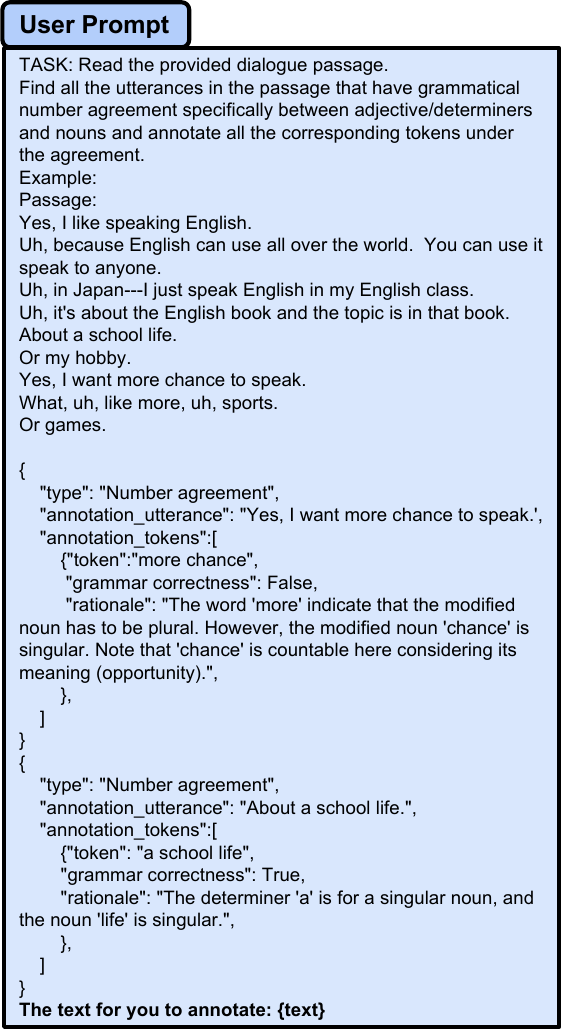}
  \vspace{-3mm}
  \caption{User prompt for annotating \textit{Numbers Agreement}. }
  \label{fig:annotation_user_prompt_numbers_agreement}
  \vspace{-5mm}
\end{figure}

We show the prompt designs used for single-sentence model outputs. Figure~\ref{fig:annotation_system_prompt} presents the system prompt with general instructions for linguistic feature extraction, while Figure~\ref{fig:annotation_user_prompt_numbers_agreement} gives the user prompt for parsing the \textit{Number Agreement} category. All prompts used in our experiments will be released in the open-source repository for the camera-ready version.



\subsection{L2 Linguistic Feature Distribution Construction and JSD/BCD Calculation}
\label{appendix:l2_feature_dist_jsd}

The construction of linguistic feature profiles follows the same procedure for both usage frequency (UF) and error rate (ER). For each model or human learner group, we first aggregate the judge annotations across the seven linguistic feature categories. For UF, we count how often each feature category appears (even if the usage is wrong). For ER, we compute the proportion of incorrect instances when they appear within each feature category.

For Jensen--Shannon divergence (JSD), both UF and ER vectors are normalized into probability distributions before comparison. That is, given a nonnegative feature vector $\mathbf{x} = (x_1, \ldots, x_K)$ over $K$ feature categories, we normalize it as
\[
p_i = \frac{x_i}{\sum_{j=1}^{K} x_j}.
\]
We then compute JSD between two normalized distributions $\mathbf{p}$ and $\mathbf{q}$:
\[
\begin{aligned}
\mathrm{JSD}(\mathbf{p}, \mathbf{q})
&= \frac{1}{2} \mathrm{KL}(\mathbf{p} \parallel \mathbf{m})
+ \frac{1}{2} \mathrm{KL}(\mathbf{q} \parallel \mathbf{m}), \\
\mathbf{m}
&= \frac{1}{2}(\mathbf{p}+\mathbf{q}).
\end{aligned}
\]

In addition to JSD, we use Bray--Curtis dissimilarity (BCD) for ER comparison. Unlike JSD, BCD is computed directly on the unnormalized ER vectors and therefore preserves differences in the raw magnitude of category-wise error tendencies. Given two nonnegative ER vectors $\mathbf{x}$ and $\mathbf{y}$, BCD is defined as
\[
\mathrm{BCD}(\mathbf{x}, \mathbf{y})
=
\frac{\sum_{i=1}^{K} |x_i-y_i|}
{\sum_{i=1}^{K} (x_i+y_i)}.
\]
A BCD value of 0 indicates identical ER profiles, while larger values indicate greater dissimilarity. Thus, JSD captures relative distributional shifts of UF and ER across feature categories, whereas BCD complements it by measuring the direct discrepancy between unnormalized ER profiles.

\subsection{Human-like L2 Linguistic Feature Assessment: Additional Analysis with Human Spoken Dialogues}
\label{appendix:l2_spoken_dialogues}

We further evaluate transfer effects using another source of human English production, ICNALE Spoken Dialogues~\cite{ishikawa_2019_icnale_spoken_dialogue}. 

Overall, the results show a trend similar to that in \ref{sec:human_likeness_assessment}. In Table~\ref{tab:jsd_human_vs_llm_combined_spoken_dialogue}, Dango ranks first in modeling human transfer effects for UF JSD, third for ER JSD, and second for ER BCD. GPT-5.5 also remains a strong prompting-based baseline.

Table~\ref{tab:jsd_l1_diff_spoken_dialogue} shows that Dango’s English production largely aligns with that of Japanese learners. English produced by Korean speakers is again closest to Dango’s production, likely because Japanese and Korean share similar word order, grammatical structures, and particle systems. We also find that the JSDs, especially ER JSD, roughly reflect the structural similarity of these languages from Japanese.

\begin{table}[t]
\centering
\small
\begin{tabular}{llll}
\toprule
\textbf{Model} &
\makecell{\textbf{JSD of}\\\textbf{UF}} &
\makecell{\textbf{JSD of}\\\textbf{ER}} &
\makecell{\textbf{BCD of}\\\textbf{ER}} \\
\midrule

\multicolumn{4}{l}{\textbf{Training-based methods}} \\
\addlinespace[2pt]
Dango
  & 0.0100\gold{}
  & 0.0214\bronze{}
  & 0.1491\silver{} \\
Unfiltered
  & 0.0113\bronze{}
  & 0.0554
  & 0.4184 \\
llm-jp-3
  & 0.0153
  & 0.0593
  & 0.7847 \\

\midrule

\multicolumn{4}{l}{\textbf{Prompting-based methods}} \\
\addlinespace[2pt]
llm-jp-3.1-instruct
  & 0.0179
  & 0.0743
  & 0.8973 \\
GPT-4o
  & 0.0140
  & 0.0104\gold{}
  & 0.3761\bronze{} \\
GPT-5.5
  & 0.0104\silver{}
  & 0.0186\silver{}
  & 0.1123\gold{} \\

\bottomrule
\end{tabular}

\caption{LLM output linguistic feature distribution divergences to human Spoken Dialogue produced by Japanese learners. Lower is more similar.}
\label{tab:jsd_human_vs_llm_combined_spoken_dialogue}
\end{table}

\begin{table}[t]
\centering
\small

\begin{tabular}{llll}
\toprule
\textbf{Human L1} &
\makecell{\textbf{JSD of}\\\textbf{UF}} &
\makecell{\textbf{JSD of}\\\textbf{ER}} &
\makecell{\textbf{BCD of}\\\textbf{ER}} \\
\midrule
Japanese   & 0.0100\gold{}   & 0.0214\silver{} & 0.1491\silver{} \\
\midrule
Korean     & 0.0136          & 0.0180\gold{}   & 0.1496\bronze{} \\
Urdu       & 0.0123\bronze{} & 0.0257          & 0.1778 \\
Mandarin   & 0.0142          & 0.0253\bronze{} & 0.1430\gold{} \\
Cantonese  & 0.0131          & 0.0436          & 0.2140 \\
Thai       & 0.0162          & 0.0303          & 0.1818 \\
Malay      & 0.0112\silver{} & 0.0404          & 0.1863 \\
\bottomrule
\end{tabular}
\caption{Linguistic feature divergences between Dango and humans with different L1 backgrounds. Human English production is from ICNALE Spoken Dialogue.}
\vspace{-3mm}
\label{tab:jsd_l1_diff_spoken_dialogue}
\end{table}


\begin{figure}
    \centering
\begin{tcolorbox}[width=\linewidth, colback=white, colframe=black!30, boxrule=0.4pt]
\scriptsize\ttfamily
\raggedright
\parindent=0pt
\begingroup
\let\n\shown
\obeylines
\begin{CJK}{UTF8}{min} 

\textbf{Japanese source}: 最終的な決定をする前に、私たちはリスクを評価する必要があります。
\textbf{GT}: We need to assess the risks before we make a final decision.

\segline

\textbf{Dango [A1--B1] on B2}:
We need to review risk before we decide the final decision.


\textbf{Correctness}: 4/5
{\color{gray}{(article omission: ``review the risk''; non-native phrasing: ``decide a decision'')}}


\scriptsize\ttfamily
\raggedright
\parindent=0pt
\let\n\shown
\obeylines

\segline

\textbf{Dango [A1--B2] on B2}:
We need to consider the risk before we make a final decision.


\textbf{Correctness}: 5/5
{\color{gray}{(comprehensible)}}

\end{CJK}

\endgroup
\end{tcolorbox}

    \caption{Case study of Dango's English production quality across training levels.}
    \label{fig:case_study_cefr_english_quality}
\end{figure}

\begin{figure}[t]
    \centering

    \includegraphics[width=0.99\linewidth]{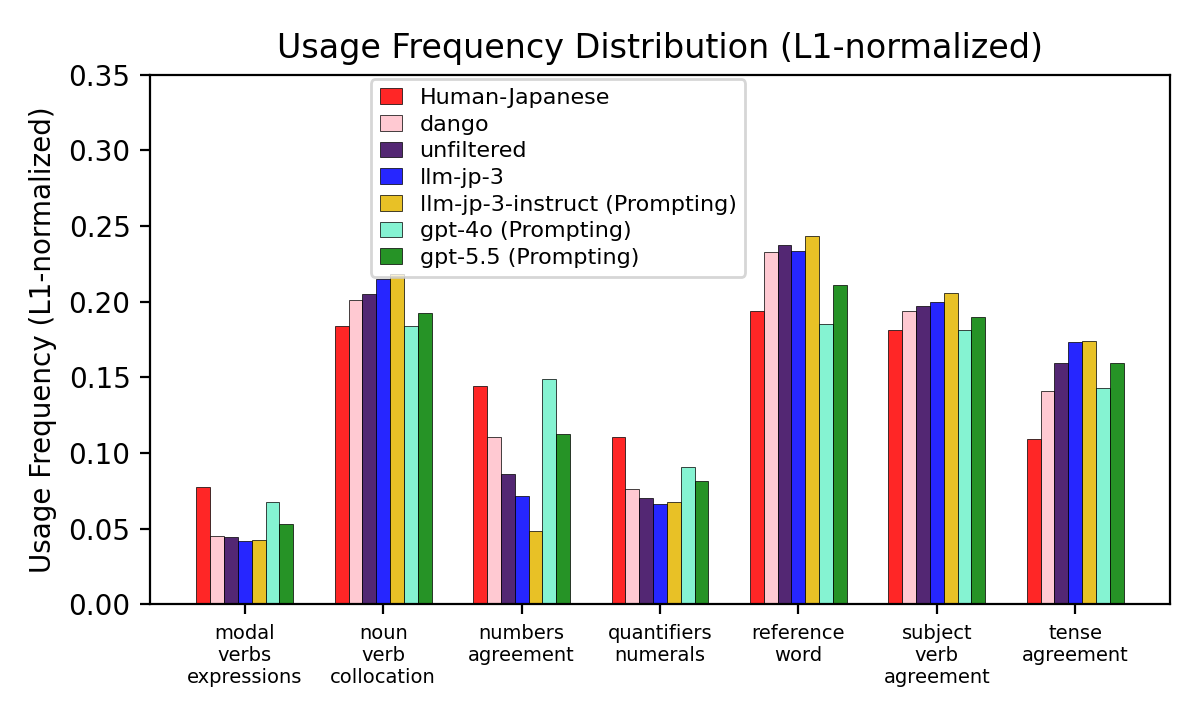}

    \vspace{-2mm}
    \caption{Usage Frequency distribution of human and LLMs for calculating JSD.}
    \label{fig:freq_dist_jsd_we}
\end{figure}

\begin{figure}[t]
    \centering

    \includegraphics[width=0.99\linewidth]{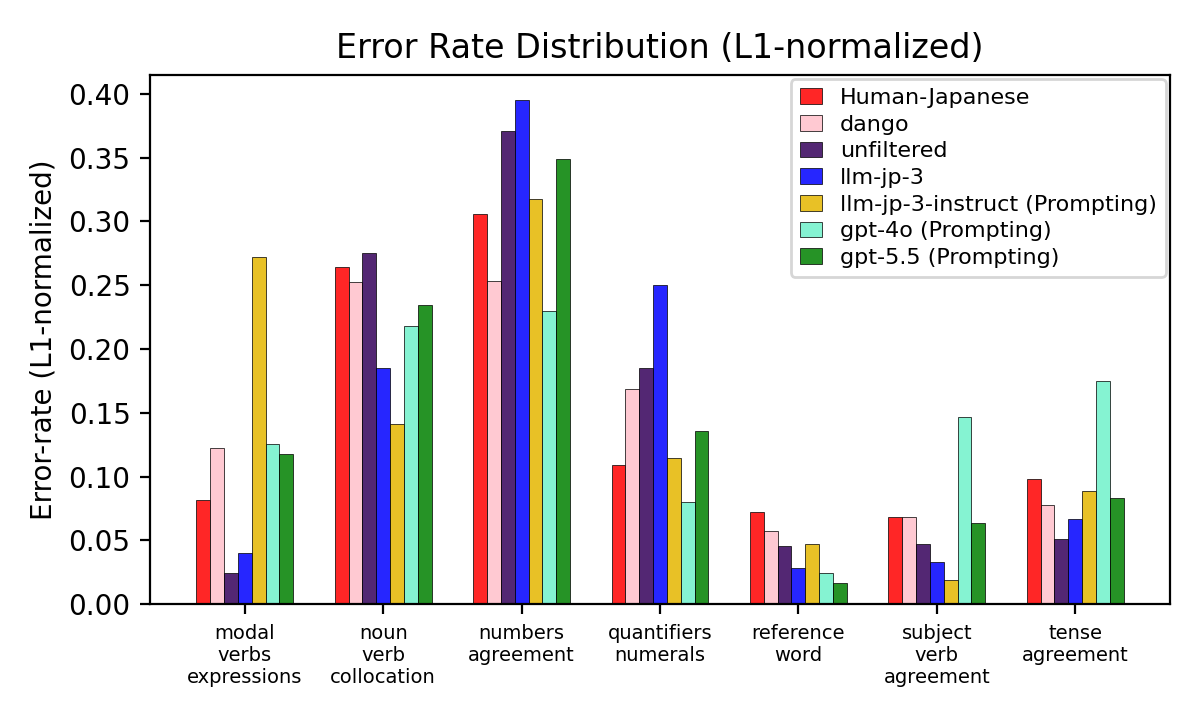}

    \vspace{-2mm}
    \caption{Error Rate distribution of human and LLMs for calculating JSD.}
    \label{fig:error_dist_jsd_we}
\end{figure}

\begin{figure}
\centering
\begin{tcolorbox}[width=\linewidth, colback=white, colframe=black!30, boxrule=0.4pt]
\begin{CJK}{UTF8}{min} 
\scriptsize\ttfamily
\raggedright
\parindent=0pt
\begingroup
\let\n\shown
\obeylines

\textbf{Japanese source}: 私は踊ることが好きです。
\textbf{Model}: I like dance.
\textbf{GT}: I like dancing.
{\color{gray}{(gerund omission: ``like dance'' instead of like dancing'')}}

\segline

\textbf{Japanese source}: 私は夕方にテレビを見ます。
\textbf{Model}: I watch TV at evening.
\textbf{GT}: I watch TV in the evening.
{\color{gray}{(preposition error: ``at evening'' instead of ``in the evening'')}}

\segline

\textbf{Japanese source}: 彼女はみんなを助けるので、天使のような人です。
\textbf{Model}: She is a angel who helps people.
\textbf{GT}: She is an angel because she helps everyone.
{\color{gray}{(article form error: ``a angel'' instead of an angel'')}}

\segline

\textbf{Japanese source}: このテレビ番組は勉強になります。
\textbf{Model}: This TV shows is studying.
\textbf{GT}: This TV show is educational.
{\color{gray}{(non-native phrasing: literal mistranslation of the Japanese idiom ``benkyou ni naru''; unnecessary plural: ``This TV shows'')}}

\segline

\textbf{Japanese source}: その犬は私の足を攻撃しました。
\textbf{Model}: The dog attack my leg.
\textbf{GT}: The dog attacked my leg.
{\color{gray}{(tense omission: ``attack'' instead of ``attacked'')}}

\endgroup
\end{CJK}
\end{tcolorbox}
\caption{Case study of grammatical and phrasing errors in Dango's English production.}
\label{fig:case_study_dango_l2_errors}
\end{figure}


\begin{figure}[t]
  \centering
  \includegraphics[width=0.97\linewidth]{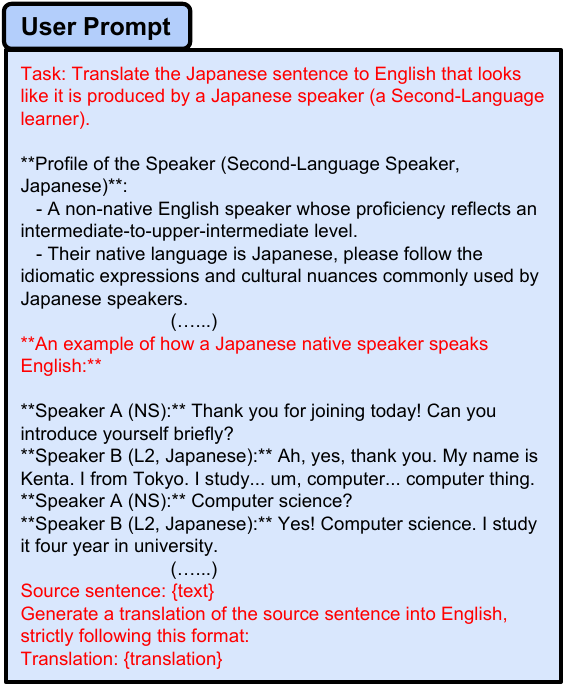}
  \vspace{-3mm}
  \caption{L1 knowledge injection user prompt for L2 role-playing. The original prompt is from \citet{gao-etal-2025-llms}; text in red indicates our modifications. }
  \label{fig:llm_l2_roleplay_user_prompt}
  \vspace{-2mm}
\end{figure}


\begin{table*}[h]
\centering
\small
\begin{tabular}{@{}llll@{}}
\toprule
\textbf{Language} & \textbf{Family} & \textbf{Order} & \textbf{Relation to Japanese (Distance)} \\ \midrule
\textbf{Japanese} & Japonic & SOV & \textit{Baseline} \\
\midrule
\textbf{Korean} & Koreanic & SOV & \textbf{Closest}: Similar syntax \& particle system \\
\textbf{Urdu} & Indo-Aryan & SOV & \textbf{Close}: Similar SOV structure \\
\textbf{Mandarin} & Sino-Tibetan & SVO & \textbf{Distant}: Different word order, partial shared script \\
\textbf{Cantonese} & Sino-Tibetan & SVO & \textbf{Distant}: Different word order, partial shared script \\
\textbf{Thai} & Kra-Dai & SVO & \textbf{Furthest}: SVO + Adjectives follow nouns \\
\textbf{Malay} & Austronesian & SVO & \textbf{Furthest}: SVO + Adjectives follow nouns \\ 
\bottomrule
\end{tabular}
\caption{Languages ranked by structural similarity to Japanese. }
\label{tab:lang_features}
\end{table*}


\begin{figure*}[t]
  \centering
  \includegraphics[width=0.95\textwidth]{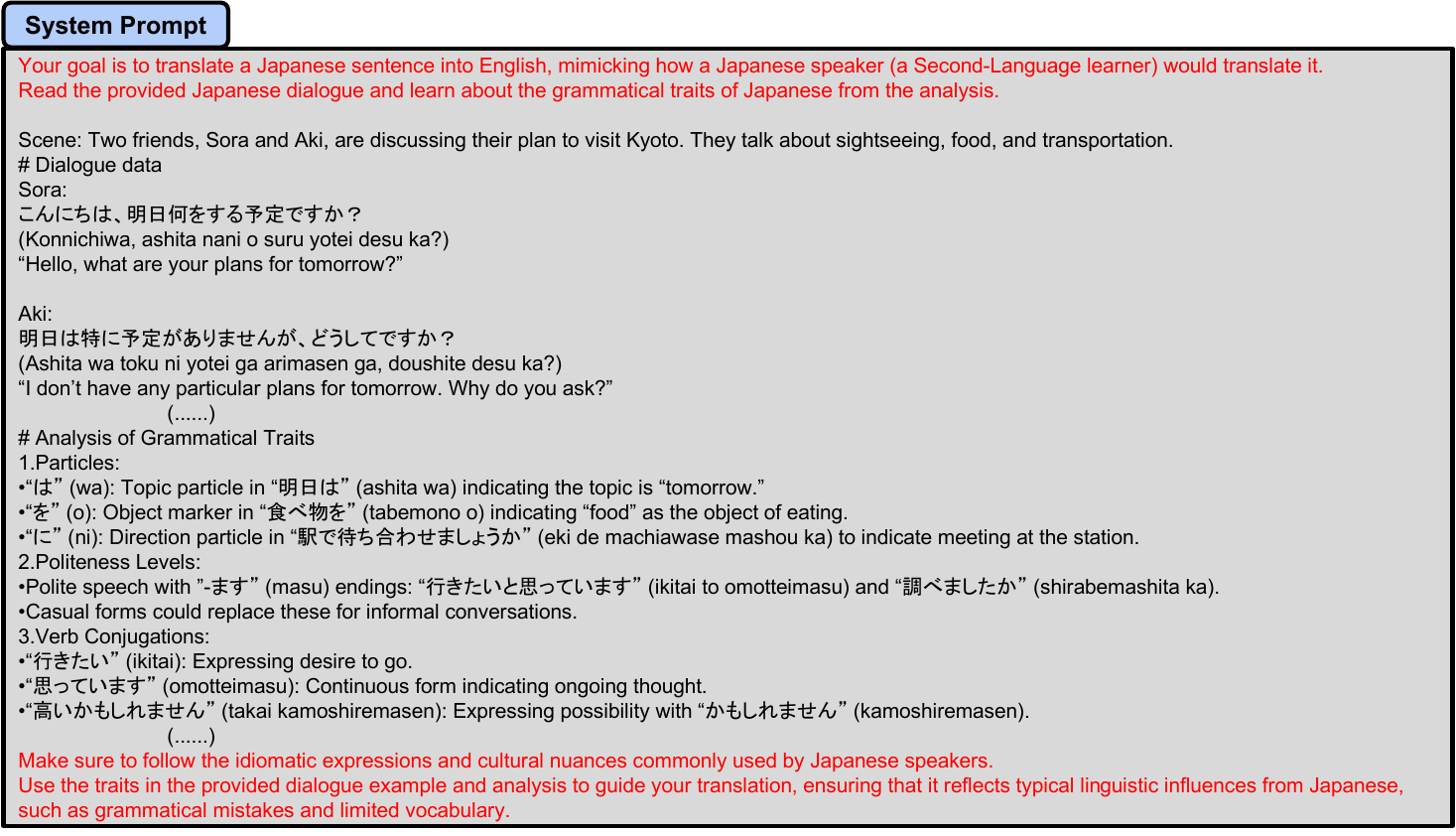}
  \vspace{-3mm}
  \caption{L1 knowledge injection system prompt for L2 role-playing. The original prompt is from \citet{gao-etal-2025-llms}; text in red indicates our modifications.}
  \label{fig:llm_l2_roleplay_system_prompt}
  \vspace{-3mm}
\end{figure*}

\begin{figure*}[t]
  \centering
  \includegraphics[width=0.95\textwidth]{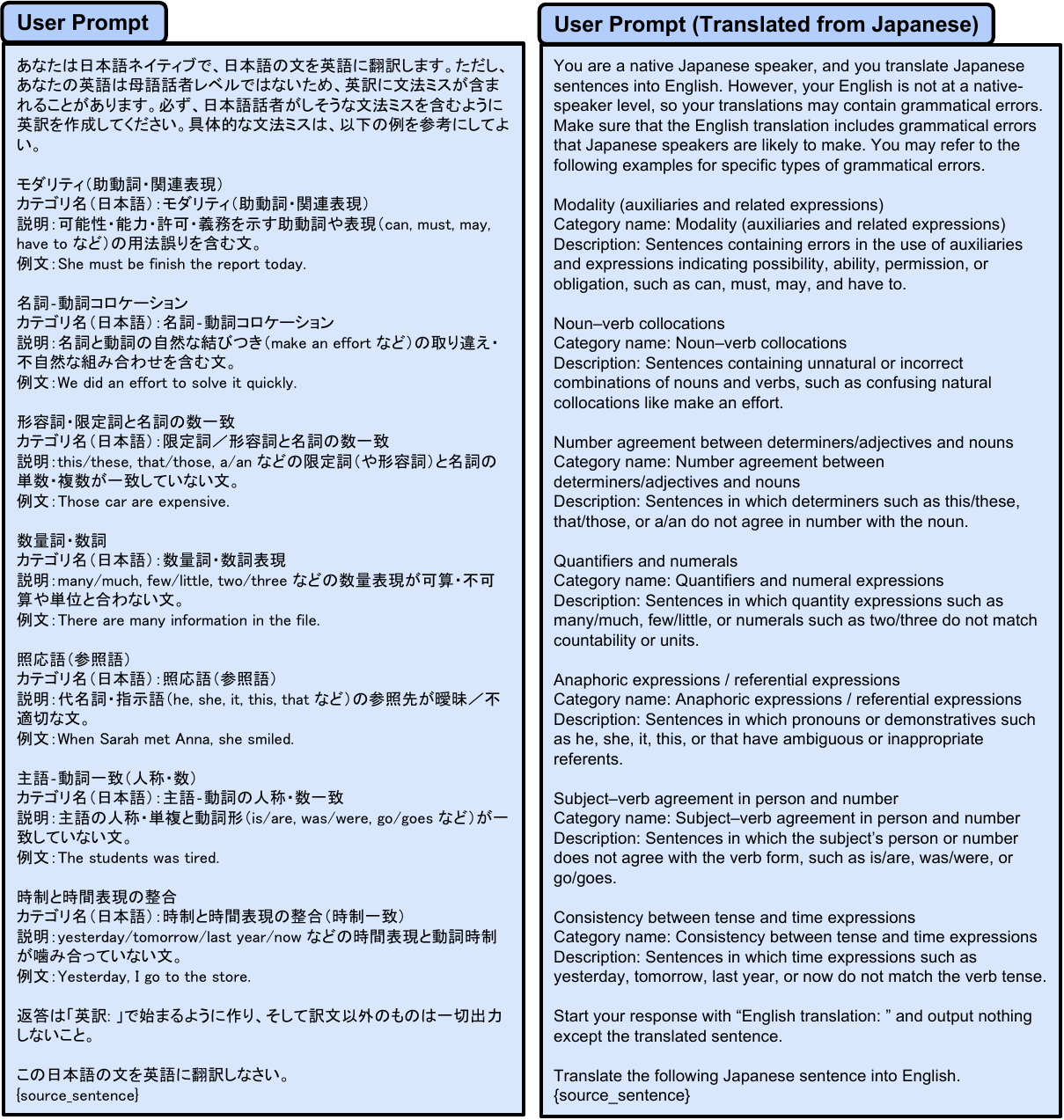}
  \vspace{-3mm}
  \caption{Our simplified L1 knowledge-injection prompt for L2 role-playing, designed for llm-jp-3.1-instruct. System prompt was not used for llm-jp-3.1-instruct. The prompt on the left was used in the experiment, while the prompt on the right is its English translation for reference. }
  \label{fig:llm-jp-3.1-instruct_l2_roleplay_user_prompt}
  \vspace{-2mm}
\end{figure*}

\begin{table*}
\centering
\small
\begin{tabular}{lrrrrr}
\toprule
\textbf{Subcategory} & \textbf{by whitelist} & \textbf{by English} & \textbf{by Chinese} & \textbf{Total Dropped} & \textbf{\# docs after filtering} \\
\midrule
CommonCrawl & 6.89\%  & 23.00\% & 0.17\% & 29.96\% & 295,309,943 \\
Kaken      & 13.13\% & 54.67\% & 0.00\% & 67.80\% & 2,228,758   \\
WARP HTML  & 2.61\%  & 13.03\% & 0.07\% & 15.70\% & 1,659,736   \\
WARP PDF   & 14.10\% & 8.61\%  & 0.18\% & 22.84\% & 35,528,902  \\
Wikipedia  & 7.35\%  & 23.82\% & 0.22\% & 31.32\% & 1,364,529   \\
\midrule
Overall    & 7.67\%  & 21.65\% & 0.17\% & 29.39\% & 336,091,868 \\
\bottomrule
\end{tabular}
\caption{Statistics of the Japanese corpus after filtering.}
\label{tab:l1_corpus_filtered_stats}
\end{table*}

\begin{table*}
\centering
\small
\begin{tabular}{l l l}
\toprule
\textbf{Filter} & \textbf{Parameter} & \textbf{Value} \\
\midrule
\multirow{2}{*}{\textbf{Unicode Whitelist}} 
 & Allowed Scripts & \begin{tabular}[c]{@{}l@{}}Japanese (Kana), CJK Ideographs, \\ ASCII, Emoji, Punctuation\end{tabular} \\
 & Document Rejection Threshold & $> 0.1\%$ non-whitelisted characters \\
\midrule
\multirow{5}{*}{\textbf{English (Latin)}} 
 & Ratio Threshold & $> 0.40$ (Latin chars / Total chars) \\
 & Max Consecutive Latin Words & $> 4$ \\
 & Max Latin Length & $> 20$ characters \\
 & Min Latin Length (Ignored) & $< 8$ characters \\
 & Document Rejection Threshold & $> 5.0\%$ violated lines \\
\midrule
\multirow{2}{*}{\textbf{Chinese}} 
 & Blocklist Source & Unihan BMP \\
 & Document Rejection Threshold & $> 0.1\%$ violated lines \\
\bottomrule
\end{tabular}
\caption{Filtering parameters and thresholds.}
\label{tab:filter_params}
\end{table*}

\end{document}